\documentclass[10pt,twocolumn,letterpaper]{article}

\usepackage[pagenumbers]{cvpr} 

\definecolor{cvprblue}{rgb}{0.21,0.49,0.74}
\usepackage[pagebackref,breaklinks,colorlinks,allcolors=cvprblue]{hyperref}

\usepackage{lipsum}
\usepackage{float}
\usepackage{multirow}
\usepackage{graphicx}
\usepackage{array}
\usepackage{makecell} 
\usepackage{supertabular}
\usepackage{booktabs}   
\usepackage{ifthen}

\definecolor{best}{HTML}{ffb2b2}
\definecolor{secondbest}{HTML}{ffd9b2}
\definecolor{thirdbest}{HTML}{ffffb2}
\newcommand{\best}[1]{{\colorbox{best}{\color{black}#1}}}
\newcommand{\sbest}[1]{{\colorbox{secondbest}{\color{black}#1}}}
\newcommand{\tbest}[1]{{\colorbox{thirdbest}{\color{black}#1}}}

\usepackage{subcaption}
\makeatletter
\renewcommand{\paragraph}{%
  \@startsection{paragraph}{4}%
  {\z@}{1.5ex \@plus 1ex \@minus .2ex}{-1em}%
  {\normalfont\normalsize\bfseries}%
}
\makeatother
\usepackage{arydshln}

\def\paperID{39942} 
\def\confName{CVPR}
\def\confYear{2026}

\title{Leveraging Multispectral Sensors for Color Correction in Mobile Cameras}

\author{
Luca Cogo$^{1}$\quad
Marco Buzzelli$^{1}$\quad
Simone Bianco$^{1}$ \quad
Javier Vazquez-Corral$^{2,3}$\quad
Raimondo Schettini$^{1}$\\
$^1$ University of Milano-Bicocca \quad $^2$ Computer Vision Center \quad $^3$ Universitat Autònoma de Barcelona\\
{\tt\small \{\href{mailto:luca.cogo@unimib.it}{\textcolor{black}{luca.cogo}}, \href{mailto:marco.buzzelli@unimib.it}{\textcolor{black}{marco.buzzelli}}, \href{mailto:simone.bianco@unimib.it}{\textcolor{black}{simone.bianco}}, \href{mailto:raimondo.schettini@unimib.it}{\textcolor{black}{raimondo.schettini}}\}@unimib.it} \\ 
{\tt\small javier.vazquez@cvc.uab.cat}
}

\begin{document}


\maketitle
\begin{abstract}

Recent advances in snapshot multispectral (MS) imaging have enabled compact, low-cost spectral sensors for consumer and mobile devices. By capturing richer spectral information than conventional RGB sensors, these systems can enhance key imaging tasks, including color correction. However, most existing methods treat the color correction pipeline in separate stages, often discarding MS data early in the process. 
We propose a unified, learning-based framework that performs end-to-end color correction and jointly leverages data from a high-resolution RGB sensor and an auxiliary low-resolution MS sensor. Our approach integrates the full pipeline within a single model, producing coherent and color-accurate outputs. 
We demonstrate the flexibility and generality of our framework by refactoring two different state-of-the-art image-to-image architectures. To support training and evaluation, we construct a dedicated dataset by aggregating and repurposing publicly available spectral datasets, rendering under multiple RGB camera sensitivities. Extensive experiments show that our approach improves color accuracy and stability, reducing error by up to 50\% compared to RGB-only and MS-driven baselines.
Code, models and dataset available at: \url{https://lucacogo.github.io/Mobile-Spectral-CC/}.

\end{abstract}

\begin{figure}
    \centering
    \includegraphics[width=1\columnwidth]{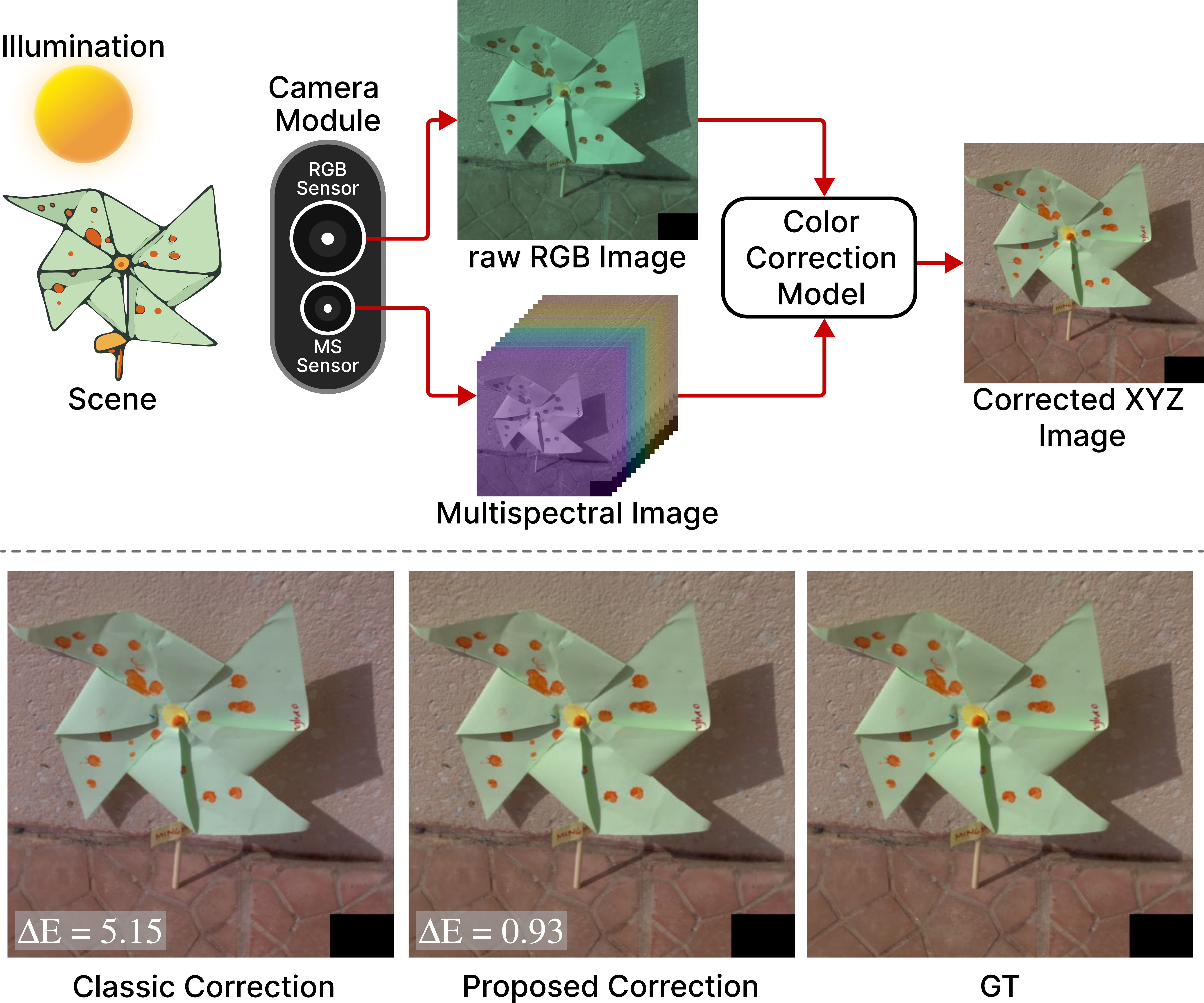}
    \caption{\textbf{Top}: Overview of the proposed color correction framework. A high-resolution RGB sensor is paired with a low-resolution multispectral sensor within a camera module. The two inputs are fused by a unified model that jointly performs illuminant estimation, illuminant discounting, and color correction to produce a color-accurate output. \textbf{Bottom}: Visual comparison between our proposed approach and a classic correction pipeline (using FC$^4$~\cite{hu2017fc4} for illuminant estimation). We report the average $\Delta$E$_{00}$ color distance and convert the images to sRGB for visualization purposes.}
    \vspace{-5mm}
    \label{fig:graphical_abstract}
\end{figure}

\vspace{-3mm}
\section{Introduction}
\vspace{-2mm}
Color correction is a fundamental component of camera imaging pipelines, responsible for transforming raw sensor measurements into perceptually accurate, device-independent color representations within standard color spaces such as CIE~XYZ~\cite{CIE19} or ProPhoto~RGB~\cite{ISO22028-2:2013}, typically under a canonical illuminant (e.g., D65). This task is particularly challenging in consumer digital photography applications, where a wide variety of lighting conditions must be handled robustly.

Traditional color correction pipelines rely on high-resolution RGB sensors and implement the process as a sequence of modular stages, namely automatic white balancing (AWB) and color space transformation (CST). 
AWB consists of illuminant estimation and illuminant discounting, which compensate for the color of the scene illumination. The subsequent transformation maps the white-balanced image from the camera’s raw color space to a device-independent representation. Although this modular approach simplifies the pipeline development, it also often propagates errors between stages, since the steps are treated independently. 
Furthermore, RGB sensors capture information in only three broad spectral channels, providing only limited spectral information about the observed scene. As a result, ambiguities arise when disentangling surface reflectance from illumination.

Recent progress in nanophotonics, semiconductor fabrication, and on-chip optics has enabled compact snapshot multispectral sensors~\cite{thomas2025trends}. In contrast to high-end laboratory-grade systems that use line-scan mechanisms or tunable filters, snapshot MS sensors utilize spectral filter arrays (SFAs), where each pixel (or small cluster of pixels) is coated with a distinct spectral filter. Although the former offer high spectral resolution, they come with drawbacks such as mechanical complexity, bulky design, and lengthy acquisition times~\cite{vagni2007survey, lapray2014multispectral}. In comparison, SFAs enable the simultaneous capture of multiple spectral bands within a single exposure. As a result, snapshot MS sensors are compact, cost-effective, and capable of real-time operation, prompting the community to explore their use in computational photography applications~\cite{zhou2024joint, tian2023enhancing, su2021multi}. Despite their lower spatial resolution and coarser spectral sampling compared to high-end devices, they provide richer spectral cues than RGB sensors, which can be leveraged to improve the color correction process.
The benefit of incorporating multispectral information into the color correction pipeline can be understood through the image formation model:
\begin{equation}
    \label{eq:image_formation}
    I_c = \int_{\lambda_{min}}^{\lambda_{max}} R(\lambda) \, E(\lambda) \, S_c(\lambda) \, d\lambda,
\end{equation}
where \(I_c\) denotes the sensor response in channel \(c\), \(R(\lambda)\) is the surface reflectance, \(E(\lambda)\) is the spectral power distribution of the illuminant, and \(S_c(\lambda)\) is the spectral sensitivity of the channel \(c\) of the camera. Accurate color correction requires disentangling \(R(\lambda)\) and \(E(\lambda)\) to achieve illuminant-invariant and device-independent color representations. Since RGB sensors provide only three measurements of this integral, the problem is underdetermined, resulting in the ambiguities described earlier. By sampling the spectrum more densely, multispectral sensors provide additional constraints that enable more accurate estimation of both \(R(\lambda)\) and \(E(\lambda)\), supporting more reliable illuminant estimation, illuminant discounting, and color space transformation.
However, most existing research, both in RGB-only and MS-driven settings, still treats these stages as independent~\cite{hu2017fc4, barron2015convolutional, barron2017fast, bianco2015color, bianco2019quasi}. In many multispectral-based pipelines, the spectral data is used only for illuminant estimation and then discarded, thereby ignoring valuable information that could enhance subsequent processing steps~\cite{koskinen2024single, glatt2024beyond}. Recent works that attempt to jointly model the entire color correction process~\cite{li2025multi} perform the computation directly on multispectral images, which limits their applicability to mobile platforms where the MS sensor typically has low spatial resolution and serves as an auxiliary modality.

In this work, we introduce a unified, learning-based framework that jointly performs illuminant estimation, illuminant discounting, and color space transformation by fusing high-resolution RGB imagery with auxiliary low-resolution MS measurements. As illustrated in \Cref{fig:graphical_abstract}, rather than using spectral cues only for illuminant estimation, our method integrates them throughout the entire pipeline, producing more stable and color-accurate results. The framework is designed for mobile configurations where a compact MS sensor complements a primary high-resolution RGB camera.
Our contributions are threefold:
\begin{itemize}
    \item We propose a dual-input, end-to-end color correction framework that fuses high-resolution RGB and low-resolution MS inputs, jointly modeling all stages of the color correction process.
    \item We construct a physically grounded dataset comprising 116,688 RGB–MS image pairs with ground-truth color references, covering a wide range of illuminants and camera sensitivities.
    \item We refactor two lightweight image-to-image architectures~\cite{conde2023perceptual, nikonorov2025color} into our framework, demonstrating the flexibility of our approach. 
\end{itemize}

\section{Related Work}
\label{sec:related}

\paragraph{Illuminant Estimation}
\label{sec:illuminant_estimation}

Illuminant estimation is the first step of automatic white balancing and aims to infer the color of the scene illumination from the captured image. Classical statistical methods rely on global color assumptions such as the Gray-World, White-Patch, and Gray-Edge hypotheses~\cite{van2005color}, which assume that the illuminant color can be inferred from simple image statistics (e.g., mean or max). These simple approaches are computationally inexpensive but often fail when their assumptions do not hold. Convolutional neural network based methods improve robustness by predicting the illuminant from learned image features; representative examples include the works of Barron et al.~\cite{barron2015convolutional,barron2017fast}, Hu et al.~\cite{hu2017fc4}, and Bianco et al.~\cite{,bianco2015color,bianco2019quasi}. 
Recent research has tackled this task using different strategies, such as learning post-capture adjustments from rendered sRGB images~\cite{afifi2019color, afifi2020deep, afifi2022auto, serrano2025revisiting} or exploiting additional scene information~\cite{afifi2025time}. In this context, the use of multispectral data has been increasingly investigated for improving illuminant estimation~\cite{khan2017illuminant, erba2024rgb, Erba2024ImprovingRI, koskinen2024single, glatt2024beyond}.
However, most existing methods rely solely on MS data rather than fusing it with RGB information, and typically discard the MS cues after illuminant estimation instead of leveraging them in subsequent processing stages. 
In this work, we address this limitation by introducing a unified RGB–MS framework that exploits multispectral information beyond illuminant estimation to achieve more accurate and consistent color correction.

\paragraph{Illuminant Discounting}
\label{sec:illuminant_discount}

After the illuminant is estimated, its influence must be removed to recover the scene reflectance. This illuminant discounting stage is commonly implemented using the von Kries adaptation model~\cite{fairchild2013color}, which applies a channel wise scaling that assumes the illuminant contribution can be neutralized independently in each channel. This yields a diagonal correction matrix that enforces a neutral response under a canonical white light. Although the von Kries model is efficient and widely used, it is known to be suboptimal, as it generally achieves a neutral appearance for gray regions but introduces noticeable chromatic errors for non-neutral surfaces~\cite{cheng2015beyond}. Several works have proposed more advanced correction strategies that integrate spatial or contextual information~\cite{finlayson1993diagonal,cheng2015beyond}, yet these remain constrained by the limited spectral information available in RGB data. Spectral measurements from multispectral sensors enable a more physically grounded separation of reflectance and illumination and directly address this limitation. In this work, given the end-to-end nature of our framework, we incorporate multispectral cues throughout the discounting process so that the model can perform more accurate illumination discount.

\paragraph{Color Space Transformation}
\label{sec:cst}

Color space transformation converts white-balanced camera RGB values into a device-independent color representation such as CIE~XYZ~\cite{CIE19} or ProPhoto~RGB~\cite{ISO22028-2:2013}. Conventional pipelines adopt pre-calibrated $3\times3$ CST matrices under multiple reference illuminants using standard color targets such as the Macbeth ColorChecker~\cite{mccamy1976color}. At runtime the estimated correlated color temperature determines interpolation weights between them, yielding the final transformation. This approach effectively normalizes the camera response but does not explicitly model interactions between illumination and reflectance, which limits color accuracy under varying lighting conditions. Other studies replace the linear transform with non-linear mappings learned end-to-end from data~\cite{hong2001study,finlayson2015color}. These methods can improve perceptual fidelity but they still rely solely on RGB inputs, which inherently constrains their accuracy. In contrast, the end-to-end formulation of our framework leverages multispectral information to inform the color space transformation (CST) directly, resulting in more perceptually accurate colors.

\paragraph{Traditional Camera Color Correction Pipelines}
\label{sec:traditional_pipeline}

Conventional camera pipelines perform color correction in two stages: automatic white balancing (AWB) and color space transformation (CST), both derived from the image formation model in \Cref{eq:image_formation}. The AWB stage discounts for the estimated illuminant using the von~Kries adaptation model~\cite{fairchild2013color}, which applies channel-wise scaling to enforce neutrality under a canonical white light.
The CST stage then maps the white-balanced RGB values to a device-independent color space such as CIE~XYZ~\cite{CIE19}. Two $3\times3$ CST matrices are typically calibrated under reference illuminants at 2500°K and 6500°K using Macbeth ColorChecker targets~\cite{mccamy1976color}, and the final transformation is obtained by interpolating between them based on the estimated correlated color temperature.
This modular pipeline, though efficient, treats illuminant estimation, discounting, and transformation as independent steps, which limits color accuracy under complex lighting. Nevertheless, it represents the standard color correction structure adopted in commercial cameras. For a fair and realistic comparison, we employ this pipeline (hereafter referred to as the traditional correction pipeline) when evaluating state-of-the-art illuminant estimation methods, as they operate solely on the first stage of the color correction pipeline. In contrast, our framework learns all stages jointly, guided by both RGB and multispectral cues.

\begin{figure*}[ht]
\centering
    \begin{subfigure}{1\linewidth}
        \includegraphics[width=\linewidth]{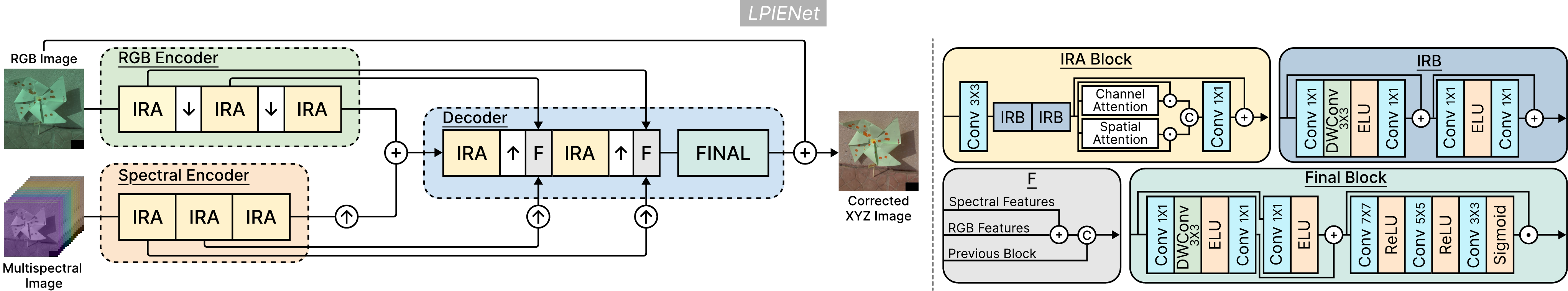} 
        \label{subfig:lpienet}
        \vspace{-3mm}
    \end{subfigure}
    \begin{subfigure}{1\linewidth}
        \includegraphics[width=\linewidth]{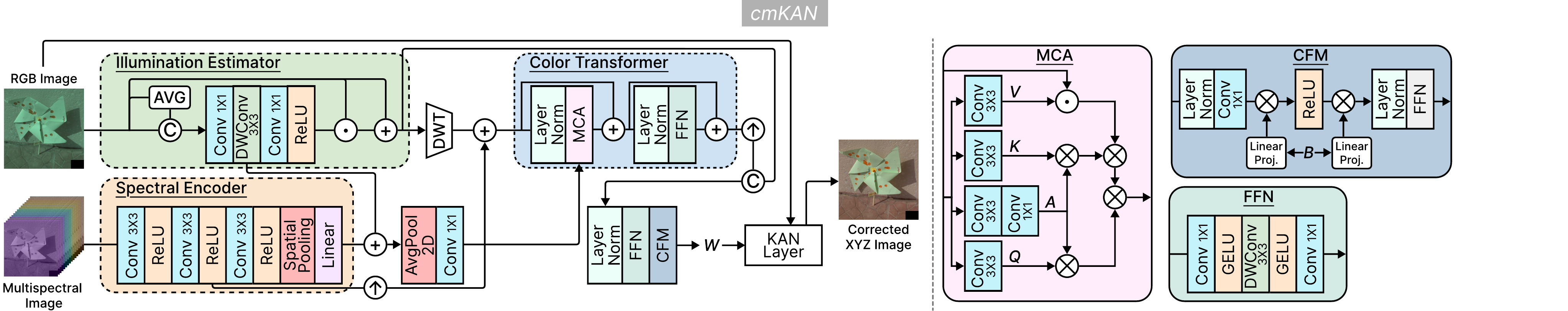}
        \label{subfig:cmkan}
        \vspace{-3mm}    
    \end{subfigure}
    \begin{subfigure}{0.65\linewidth}
        \includegraphics[width=\linewidth]{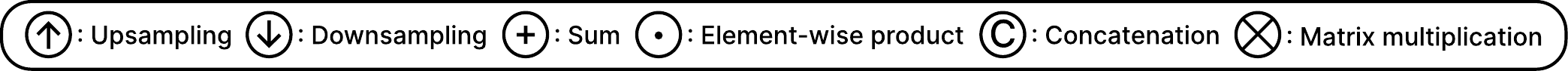} 
    \end{subfigure}
\caption{Overview of the proposed architectures adapted from (\textbf{top}) LPIENet~\cite{conde2023perceptual} and (\textbf{bottom}) cmKAN~\cite{nikonorov2025color}. For both architectures, we add a spectral encoder module and fuse the features with the ones extracted from the RGB image. For visualization purposes the output and the input RGB images are respectively converted to sRGB and gamma corrected.}
\vspace{-3mm}    
\label{fig:methods}
\end{figure*}

\section{Method}
\label{sec:method}

We aim at performing the full color correction pipeline by leveraging both a high-resolution RGB image and an auxiliary low-resolution multispectral capture. The two modalities are fused within an image-to-image framework that implicitly estimates the scene illuminant, compensates for both its color cast and the spectral sensitivity of the camera, and produces color-corrected outputs directly in the CIE~XYZ color space. We adapt to this setting two recent lightweight architectures, LPIENet~\cite{conde2023perceptual} and cmKAN~\cite{nikonorov2025color}, modifying their designs to process the additional MS input. \Cref{fig:methods} shows a schematic of both architectures.

\paragraph{LPIENet}
was originally designed for general-purpose image restoration and optimized for lightweight deployment on mobile hardware. Its structure follows a U-Net-like topology~\cite{ronneberger2015u}, which makes it a suitable foundation for image-to-image problems requiring spatially coherent transformations.
The core component of the architecture is the IRA block (Inverted Residual Attention), which integrates convolutional blocks inspired by the MobileNet architecture~\cite{howard2017mobilenets} with parallel channel and spatial attention mechanisms~\cite{woo2018cbam}.
The network is composed of six main blocks: three encoder blocks and two decoder blocks, connected through skip connections, followed by a final refinement block.
To integrate the multispectral modality, we introduce an additional spectral encoder that mirrors the original encoder architecture and consists of three IRA blocks without downsampling.
The features extracted from each encoder of the multispectral branch are fused with those from the RGB encoder through the skip connections, where element-wise addition is applied before each decoding stage. The decoder structure is maintained identical to the original LPIENet.
We conduct experiments using two model configurations. The first follows the original feature sizes \([16, 32, 64, 32, 16]\) across the encoder-decoder blocks, while the second one uses a ``small" configuration \([8, 16, 32, 16, 8]\). Both models are highly lightweight, containing approximately 220K and 60K trainable parameters, respectively.

\vspace{1mm}

\paragraph{cmKAN}
is an image-to-image architecture that differs from conventional encoder–decoder designs. Originally designed for color matching, it employs a hypernetwork generator that predicts spatially varying parameters controlling the splines of a Kolmogorov-Arnold Network (KAN) layer~\cite{liu2024kan}, enabling smooth and non-linear color transformations across the image. 
In this work, we refactor the lightweight generator variant proposed by the original authors, which comprises three principal modules: the Illumination Estimator (IE), which captures illumination information; the Color Transformer (CT), which incorporates a Multi-Scale Color Attention (MCA) mechanism to encode inter-channel dependencies; and the Color Feature Modulator (CFM), which fuses the learned features and projects them into the KAN parameter space. 
Consistent with the original implementation, a single KAN layer is employed, using B-splines of order $k=3$ and a grid size $G=5$. 
To integrate multispectral information, a compact spectral encoder composed of three convolutional layers with $3\times3$ kernels and ReLU activations is introduced. 
Spectral features are incorporated within the generator at two distinct feature levels through element-wise addition with the outputs produced by the Illumination Estimator block. The entire architecture is highly compact, with only 18K trainable parameters in total.

\begin{figure*}[ht]
\centering
    \begin{subfigure}{0.55\linewidth}
        \includegraphics[width=\linewidth]{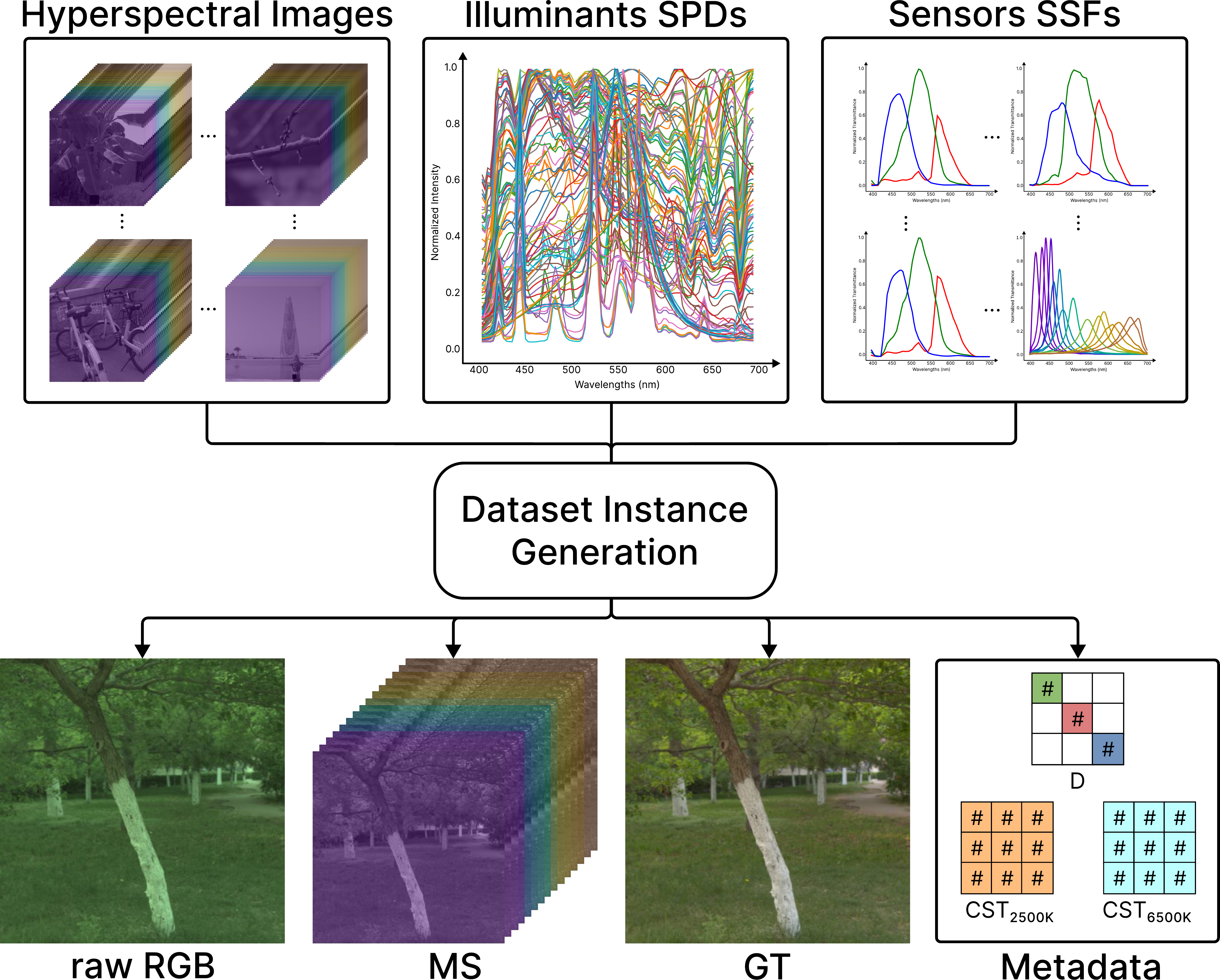} 
        \caption{}
    \end{subfigure}
    \begin{subfigure}{0.442\linewidth}
        \includegraphics[width=\linewidth]{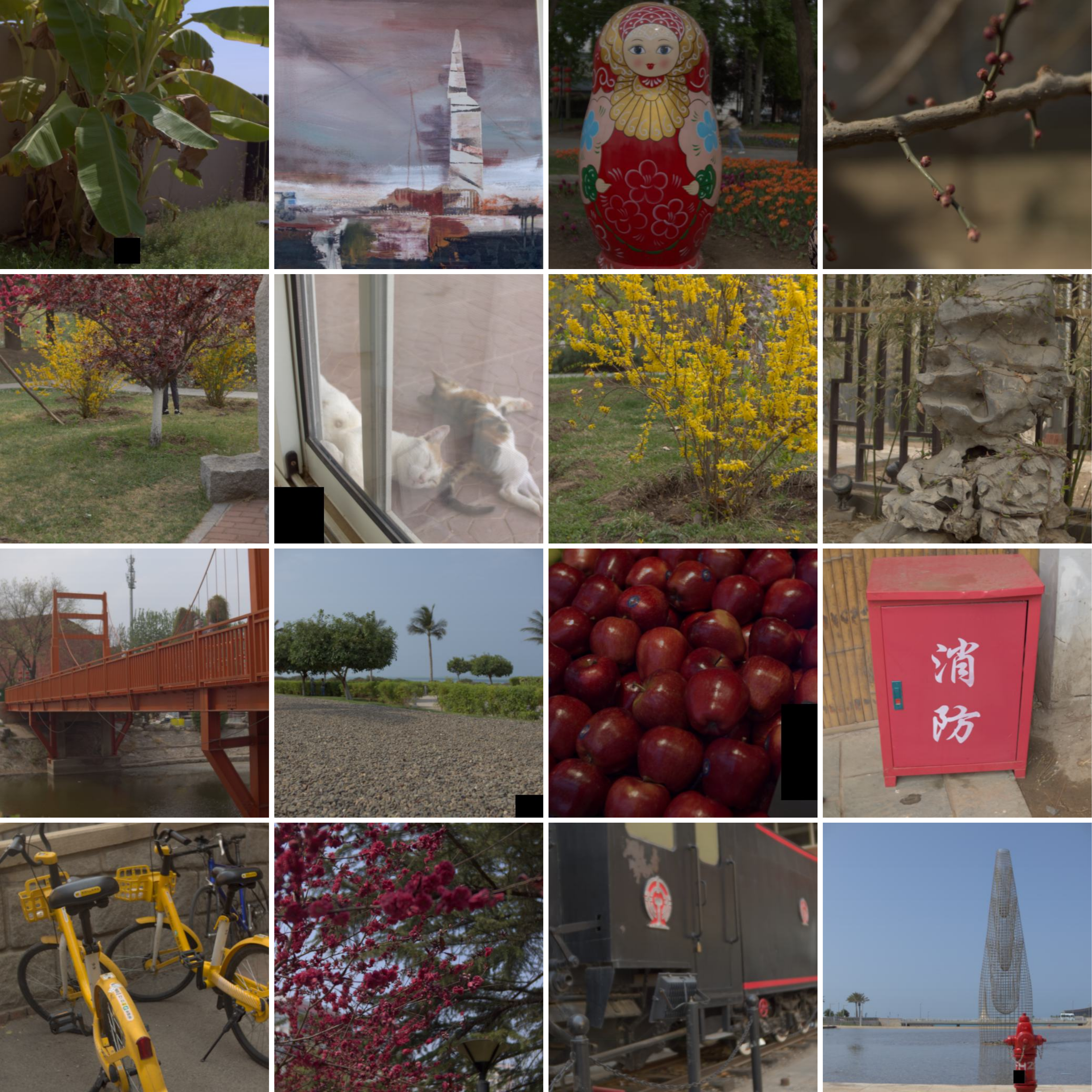}
        \caption{}
    \end{subfigure}
\vspace{-4mm}  
\caption{(a) Overview of the dataset generation pipeline. Hyperspectral reflectance images are rendered using multiple camera sensitivities under different illuminants with known spectral power distributions (SPDs) to produce RGB, MS and GT image triplets. (b) Representative samples of the proposed dataset. For visualization purposes, images are converted to sRGB.}
\label{fig:dataset}
\end{figure*}

\section{Dataset}
\label{sec:dataset}
To validate the proposed framework, we require a dataset comprising high-resolution RGB and low-resolution MS acquisitions, along with ground-truth images in the CIE~XYZ color space under a canonical illuminant (D65). Such data allow quantitative evaluation of the color correction process within a physically consistent environment.
Existing datasets for color constancy and color correction~\cite{gehler2008bayesian,cheng2014illuminant,ershov2020cube++} or for RGB+MS imaging~\cite{glatt2024beyond} lack the spectral and colorimetric information necessary to derive ground-truth reflectance and illumination. In practice, accurate color ground-truths cannot be obtained from RGB or MS captures alone, since this requires hyperspectral reflectance measurements that permit re-illumination under a known illuminant. Consequently, no real dataset currently satisfies the requirements of our task.
To overcome this limitation, we construct a physically grounded simulated dataset derived from two publicly available hyperspectral datasets with densely sampled spectral reflectance images~\cite{li2021multispectral,du2025automatic}. Because these scenes were captured with real hyperspectral cameras, they inherently include realistic sensor noise, optical blur, and acquisition artifacts. Using these reflectance spectra, we synthesize corresponding RGB and MS measurements according to the image formation model in \Cref{eq:image_formation} and render ground-truth color references under the canonical D65 illuminant.
\Cref{fig:dataset} illustrates the dataset generation procedure and some representative scenes. Each scene is re-illuminated under 102 illuminants from the dataset of Barnard et al.~\cite{barnard2002data}, providing a wide variety of lighting conditions. RGB images are simulated using several representative camera spectral sensitivities from mobile devices\footnote{Google~Pixel~3, iPhone~Xs~Max, Huawei~Mate~20~Pro, Samsung~Galaxy~Note~9} and mirrorless cameras\footnote{Canon~R5, Sony~$\alpha$9~III, Nikon~Zf}, while the MS images are rendered using the sensitivities of a Spectricity S1 sensor~\cite{spectricity}, comprising 15 narrowband channels across the visible range. RGB images are rendered at $512\times512$ pixels, matching the native resolution of the hyperspectral data, and MS images at $64\times64$ pixels to emulate a compact snapshot spectral sensor.
Prior to rendering, we mask the white reference patches used for calibration and perform manual quality control to remove corrupted or incomplete data. After filtering, 1,144 valid scenes remain. Combined with the 102 illuminants, this results in 116,688 image triplets of high-resolution RGB, low-resolution MS, and CIE~XYZ ground-truth under D65 illuminant.  In addition, we compute metadata for each image, including the ground-truth illuminant RGB triplet and the color space transformation (CST) presets required for the traditional color correction pipeline.
To prevent data leakage, we apply a scene-wise split so that test and validation scenes remain entirely unseen during training. Specifically, 80\% of scenes are used for training and 20\% for testing, with 20\% of the training subset reserved for validation.

\paragraph{Simulating Spatial Misalignments}
To emulate geometric inconsistencies in dual-sensor systems, we generate a misaligned version of the aforementioned dataset by introducing spatial offsets to the RGB–MS pairs. Realistic warping transformations are derived from the Zurich dataset~\cite{ignatov2020replacing}, which provides 168 real misaligned image pairs. For each simulated scene, a homography randomly sampled from this distribution is applied to the MS image, introducing plausible misalignment while preserving physical consistency. The misaligned dataset mirrors the structure and partitioning of the aligned version.

\begin{table*}[ht]
\caption{Results on the aligned version of the proposed dataset, aggregated by camera type: mirrorless (Canon~R5, Nikon~Zf, Sony~$\alpha$9~III) and mobile (Google~Pixel~3, Huawei~Mate~20~Pro, iPhone~Xs~Max, Samsung~Galaxy~Note~9). Aggregation is performed by averaging the results of each camera. We highlight \best{best}, \sbest{second best} and \tbest{third best} results for each metric. Per-camera results are provided in the Supplementary Material.}
\label{tab:main_results}
\resizebox{\linewidth}{!}{
\begin{tabular}{llcccccccccccccccc}
    \toprule
                                            &                                           & \multicolumn{8}{c}{$\Delta$E$_{00}$ $\downarrow$}   & \multicolumn{8}{c}{Reproduction Error $\downarrow$}                 \\ \cmidrule(lr){3-10} \cmidrule(lr){11-18} \\\\\noalign{\vskip -2.25em}
    Camera                                  & Method                                    & Mean                  & Med.             & Tri.                & B-25              & W-25                  & 95-P                  & 99-P                  & Max          & Mean                  & Med.             & Tri.                & B-25              & W-25                  & 95-P                  & 99-P                  & Max               \\ \midrule
    \multirow{15}{*}{\vtop{\hbox{\strut Mirrorless}\hbox{\strut Sensors}}} 

                                     & GW~\cite{van2005color}                & 7.80 & 7.15 & 7.28 & 2.81 & 13.63 & 15.92 & 21.89 & 26.15 & 9.42 & 8.07 & 8.24 & 2.75 & 18.91 & 23.89 & 33.06 & 42.69 \\
                                     & WP~\cite{van2005color}                 & 5.08 & 4.23 & 4.57 & 1.89 & 9.62  & 11.09 & 14.60 & 22.60 & 5.93 & 4.47 & 4.90 & 1.43 & 12.73 & 15.64 & 20.78 & 38.99 \\
                                     & GGW~\cite{van2005color}               & 6.13 & 5.47 & 5.61 & 2.05 & 11.53 & 13.65 & 19.49 & 24.70 & 7.31 & 5.88 & 5.98 & 1.79 & 15.51 & 19.82 & 29.29 & 40.85 \\
                                     & SoG~\cite{van2005color}               & 5.94 & 5.32 & 5.43 & 2.07 & 11.08 & 13.35 & 17.75 & 30.62 & 7.06 & 5.70 & 5.77 & 1.70 & 15.02 & 19.01 & 26.19 & 45.11 \\
                                     & GE1~\cite{van2005color}               & 5.53 & 4.91 & 4.92 & 1.84 & 10.54 & 12.72 & 18.79 & 24.90 & 6.40 & 4.79 & 5.08 & 1.47 & 14.03 & 17.65 & 22.60 & 40.34 \\
                                     & GE2~\cite{van2005color}               & 5.49 & 5.05 & 5.04 & 1.99 & 9.99  & 12.17 & 14.82 & 25.03 & 6.35 & 5.00 & 5.29 & 1.49 & 13.66 & 16.58 & 22.37 & 56.92 \\
                                     & Squeezenet-FC$^4$~\cite{hu2017fc4} & 3.28 & 2.86 & 2.97 & 1.45 & 5.82  & 6.72  & 9.90  & \sbest{18.64} & 4.38 & 2.78 & 3.17 & 0.92 & 10.57 & 13.68 & 18.37 & \sbest{37.72} \\
                                     & ConvMean~\cite{Gong2019ConvolutionalMA}          & 3.55 & 2.98 & 3.09 & 1.43 & 6.67  & 7.89  & 12.34 & 20.58 & 4.61 & 2.91 & 3.35 & 0.95 & 11.13 & 13.90 & 19.58 & 41.24 \\
                                     & QU~\cite{bianco2019quasi}             & 6.37 & 5.73 & 5.79 & 2.31 & 11.70 & 13.90 & 18.84 & 26.52 & 7.88 & 5.88 & 6.36 & 2.03 & 17.04 & 21.42 & 29.43 & 41.75 \\
                                     & QU+ft~\cite{bianco2019quasi}             & 5.30 & 4.65 & 4.76 & 1.95 & 9.89  & 11.90 & 16.62 & 24.42 & 6.60 & 5.12 & 5.35 & 1.72 & 13.99 & 17.51 & 27.04 & 39.77 \\
                                     & SpectralFC$^4$~\cite{hu2017fc4}    & 3.25 & 2.79 & 2.88 & 1.35 & 5.98  & 6.90  & 10.66 & \tbest{20.42} & 4.31 & 2.73 & 3.14 & 0.88 & 10.45 & 13.01 & 18.74 & \best{36.91} \\
                                     & SpectralConvMean~\cite{Gong2019ConvolutionalMA}  & 3.17 & 2.73 & 2.83 & 1.33 & 5.77  & 6.56  & 10.50 & 23.40 & 4.24 & 2.70 & 3.05 & \tbest{0.87} & 10.27 & 13.03 & 18.47 & 40.77 \\ \\\noalign{\vskip -1em} \cdashline{2-18} \\\noalign{\vskip -1em}
                                     & LPIENet~(Ours)           & \sbest{1.74} & \sbest{1.55} & \sbest{1.59} & \sbest{0.99} & \best{2.80}  & \best{3.18}  & \best{4.54}  & \best{14.10} & \sbest{3.23} & \sbest{1.53} & \sbest{2.01} & \sbest{0.67} & \sbest{8.70}  & \sbest{10.78} & \sbest{17.29} & 38.49 \\
                                     & LPIENet-small~(Ours)     & \tbest{2.09} & \tbest{1.88} & \tbest{1.92} & \tbest{1.26} & \tbest{3.29}  & \tbest{3.73}  & \sbest{5.41}  & 21.10 & \tbest{3.57} & \tbest{1.91} & \tbest{2.35} & 0.88 & \tbest{9.09}  & \tbest{11.24} & \tbest{17.56} & \tbest{38.05} \\
                                     & cmKAN-light~(Ours)        & \best{1.60} & \best{1.30} & \best{1.36} & \best{0.71} & \sbest{3.02}  & \sbest{3.55}  & \tbest{5.74}  & 21.19 & \best{2.91} & \best{1.21} & \best{1.71} & \best{0.42} & \best{8.26}  & \best{10.33} & \best{16.66} & 38.21 \\ \midrule
\multirow{15}{*}{\vtop{\hbox{\strut Mobile}\hbox{\strut Sensors}}} 

                                    & GW~\cite{van2005color}                & 7.71                      & 7.06                      & 7.19                      & 2.79                      & 13.49                     & 15.92                     & 21.45                     & 25.87                      & 9.48                      & 8.16                      & 8.35                      & 2.86                      & 18.76                     & 23.31                      & 33.47                      & 42.85                      \\
                                    & WP~\cite{van2005color}                 & 5.04                      & 4.20                      & 4.50                      & 1.95                      & 9.46                      & 10.96                     & 14.84                     & \sbest{18.75}              & 6.16                      & 4.76                      & 5.10                      & 1.59                      & 13.02                     & 15.68                      & 20.83                      & 40.11                      \\
                                    & GGW~\cite{van2005color}               & 6.15                      & 5.46                      & 5.61                      & 2.09                      & 11.47                     & 13.43                     & 19.03                     & 24.22                      & 7.57                      & 6.09                      & 6.21                      & 1.94                      & 15.83                     & 20.25                      & 29.29                      & 41.54                      \\
                                    & SoG~\cite{van2005color}               & 5.96                      & 5.32                      & 5.46                      & 2.12                      & 11.06                     & 13.28                     & 17.41                     & 22.50                      & 7.33                      & 5.94                      & 6.01                      & 1.86                      & 15.39                     & 19.45                      & 27.20                      & 41.53                      \\
                                    & GE1~\cite{van2005color}               & 5.50                      & 4.87                      & 4.90                      & 1.88                      & 10.46                     & 12.71                     & 17.95                     & 21.99                      & 6.59                      & 4.98                      & 5.28                      & 1.58                      & 14.36                     & 17.81                      & 23.59                      & 41.85                      \\
                                    & GE2~\cite{van2005color}               & 5.44                      & 4.91                      & 4.97                      & 2.01                      & 9.90                      & 12.35                     & 14.91                     & 22.28                      & 6.50                      & 5.04                      & 5.36                      & 1.55                      & 13.83                     & 16.62                      & 22.29                      & 42.06                      \\
                                    & Squeezenet-FC$^4$~\cite{hu2017fc4} & 3.16                      & 2.75                      & 2.83                      & 1.40                      & 5.63                      & 6.65                      & 9.47                      & \tbest{19.76}              & 4.51                      & 2.87                      & 3.29                      & 0.98                      & 10.80                     & 13.63                      & 18.71                      & 38.39                      \\
                                    & ConvMean~\cite{Gong2019ConvolutionalMA}          & 3.54                      & 2.95                      & 3.07                      & 1.44                      & 6.68                      & 7.98                      & 12.29                     & 21.16                      & 4.76                      & 3.05                      & 3.49                      & 1.03                      & 11.40                     & 14.16                      & 19.53                      & 40.20                      \\
                                    & QU~\cite{bianco2019quasi}                & 6.37                      & 5.80                      & 5.87                      & 2.31                      & 11.65                     & 14.29                     & 19.28                     & 21.94                      & 8.10                      & 6.34                      & 6.71                      & 2.09                      & 17.17                     & 21.39                      & 30.17                      & 42.07                      \\
                                    & QU+ft~\cite{bianco2019quasi}             & 5.41                      & 4.79                      & 4.87                      & 1.97                      & 10.04                     & 11.87                     & 17.33                     & 20.87                      & 7.03                      & 5.55                      & 5.80                      & 1.85                      & 14.70                     & 18.53                      & 28.58                      & 40.69                      \\
                                    & SpectralFC$^4$~\cite{hu2017fc4}    & 3.19                      & 2.77                      & 2.85                      & 1.39                      & 5.76                      & 6.68                      & 10.29                     & 19.80                      & 4.38                      & 2.76                      & 3.17                      & 0.96                      & 10.53                     & 13.38                      & 19.05                      & \tbest{37.70}              \\
                                    & SpectralConvMean~\cite{Gong2019ConvolutionalMA}  & 3.16                      & 2.73                      & 2.82                      & 1.39                      & 5.68                      & 6.43                      & 10.45                     & 22.38                      & 4.35                      & 2.86                      & 3.21                      & 0.98                      & 10.28                     & 12.77                      & 18.76                      & 40.70                      \\ \\\noalign{\vskip -1em} \cdashline{2-18} \\\noalign{\vskip -1em}
                                    & LPIENet~(Ours)           & \sbest{1.66}              & \sbest{1.49}              & \sbest{1.53}              & \sbest{0.94}              & \best{2.69}               & \best{3.06}               & \best{4.38}               & \best{12.73}               & \sbest{3.13}              & \sbest{1.52}              & \sbest{1.96}              & \sbest{0.66}                      & \sbest{8.38}              & \best{10.24}               & \sbest{16.53}              & \tbest{37.70}              \\
                                    & LPIENet-small~(Ours)     & \tbest{1.84}              & \tbest{1.64}              & \tbest{1.68}              & \tbest{1.07}              & \tbest{2.96}              & \tbest{3.37}              & \sbest{5.00}              & 21.20                      & \tbest{3.28}              & \tbest{1.66}              & \tbest{2.11}              & \tbest{0.75}                      & \tbest{8.59}              & \tbest{10.60}              & \tbest{16.68}              & \best{37.43}               \\
                                    & cmKAN-light~(Ours)        & \best{1.47}               & \best{1.19}               & \best{1.24}               & \best{0.65}               & \sbest{2.79}              & \sbest{3.30}              & \tbest{5.61}              & 20.78                      & \best{2.83}               & \best{1.13}               & \best{1.62}               & \best{0.40}                      & \best{8.15}               & \sbest{10.25}              & \best{16.35}               & \sbest{37.56}              \\ \bottomrule
    
    \end{tabular}
}

\end{table*}

\begin{figure*}[ht]
\captionsetup[subfigure]{labelformat=empty}

\centering
    \begin{subfigure}{0.135\linewidth}
    \includegraphics[width=\linewidth]{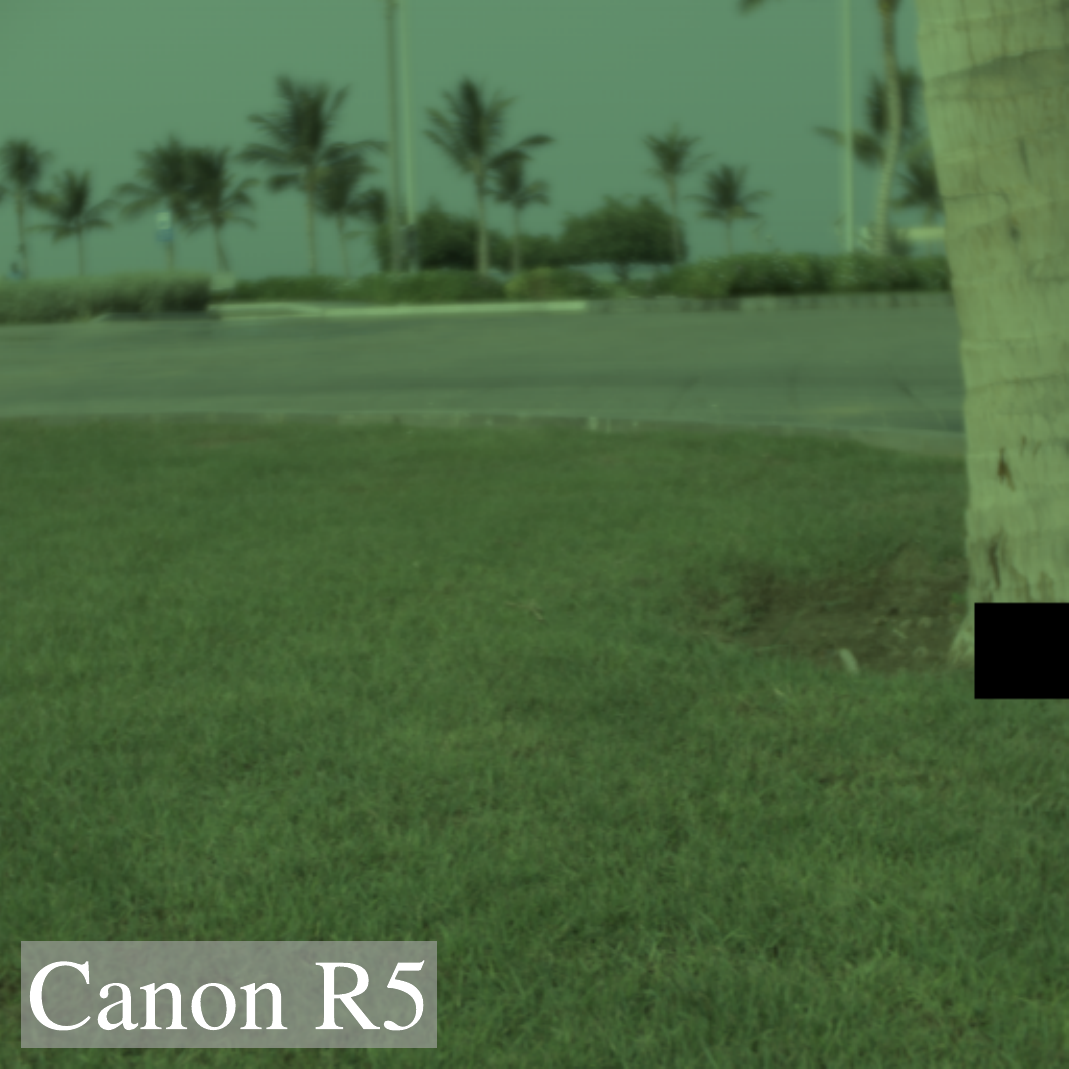} 
    \end{subfigure}
    \begin{subfigure}{0.0078\linewidth}
    \includegraphics[width=\linewidth]{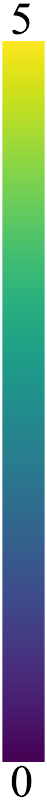} 
    \end{subfigure}
    \begin{subfigure}{0.135\linewidth}
    \includegraphics[width=\linewidth]{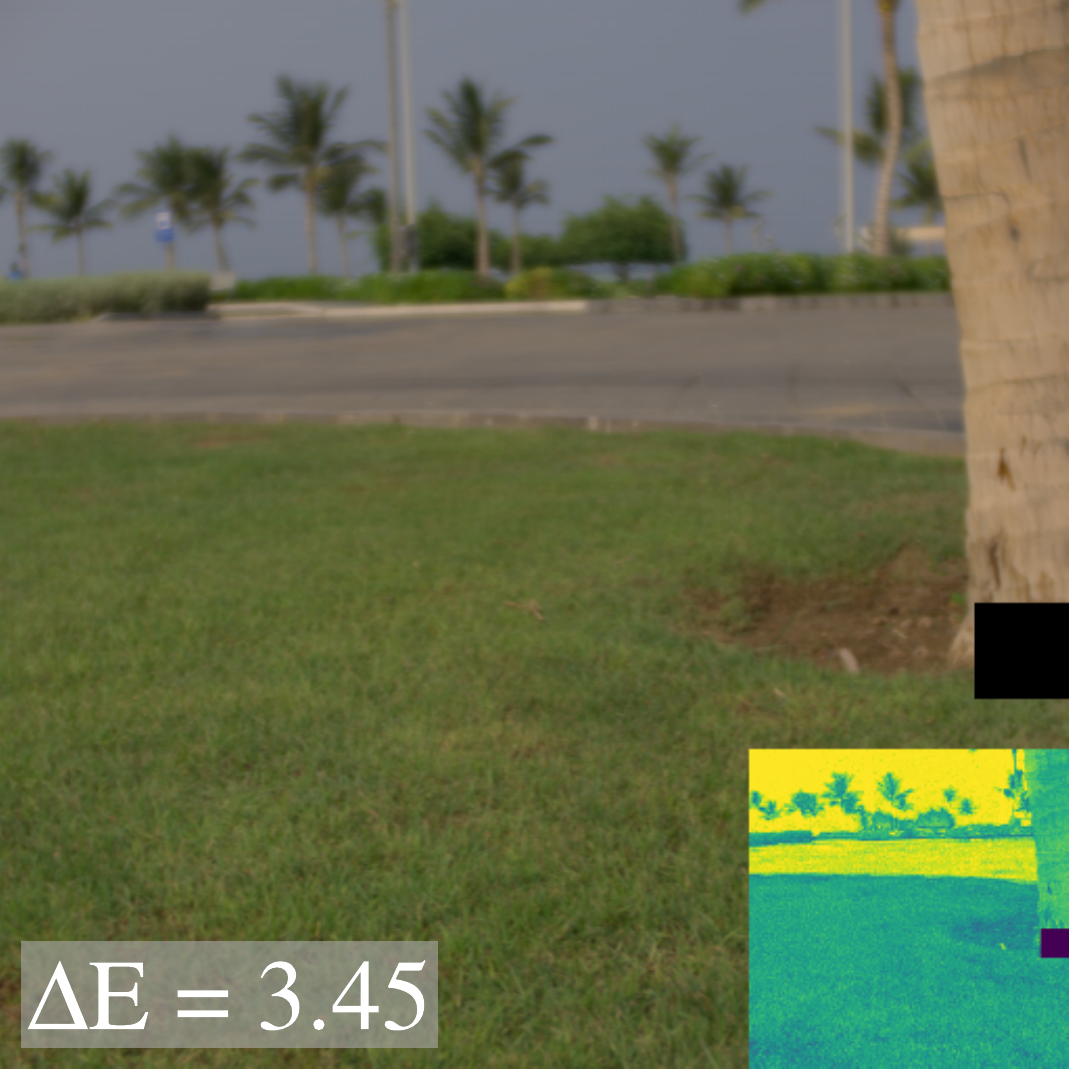} 
    \end{subfigure}
    \begin{subfigure}{0.135\linewidth}
    \includegraphics[width=\linewidth]{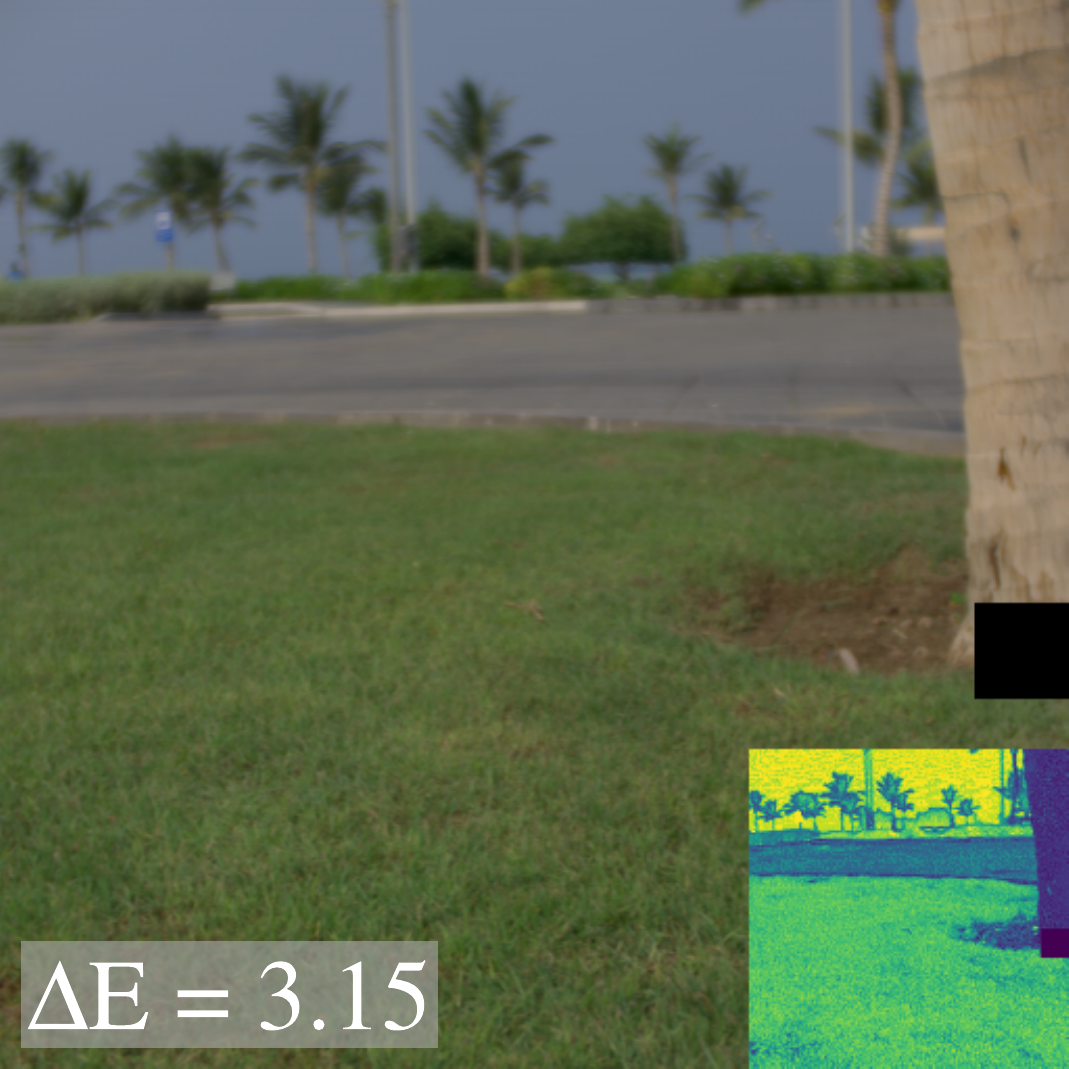} 
    \end{subfigure}
    \begin{subfigure}{0.135\linewidth}
    \includegraphics[width=\linewidth]{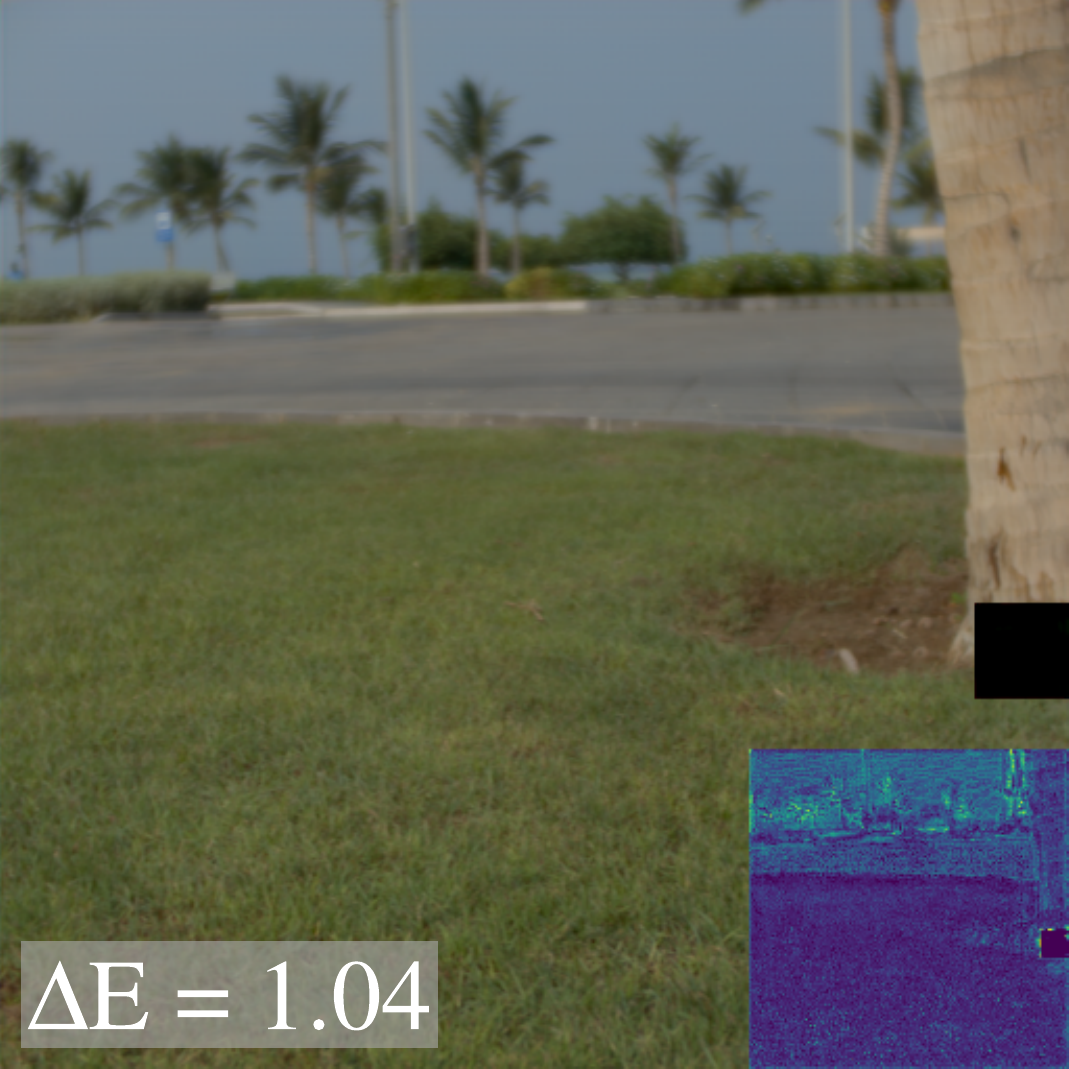} 
    \end{subfigure}
    \begin{subfigure}{0.135\linewidth}
    \includegraphics[width=\linewidth]{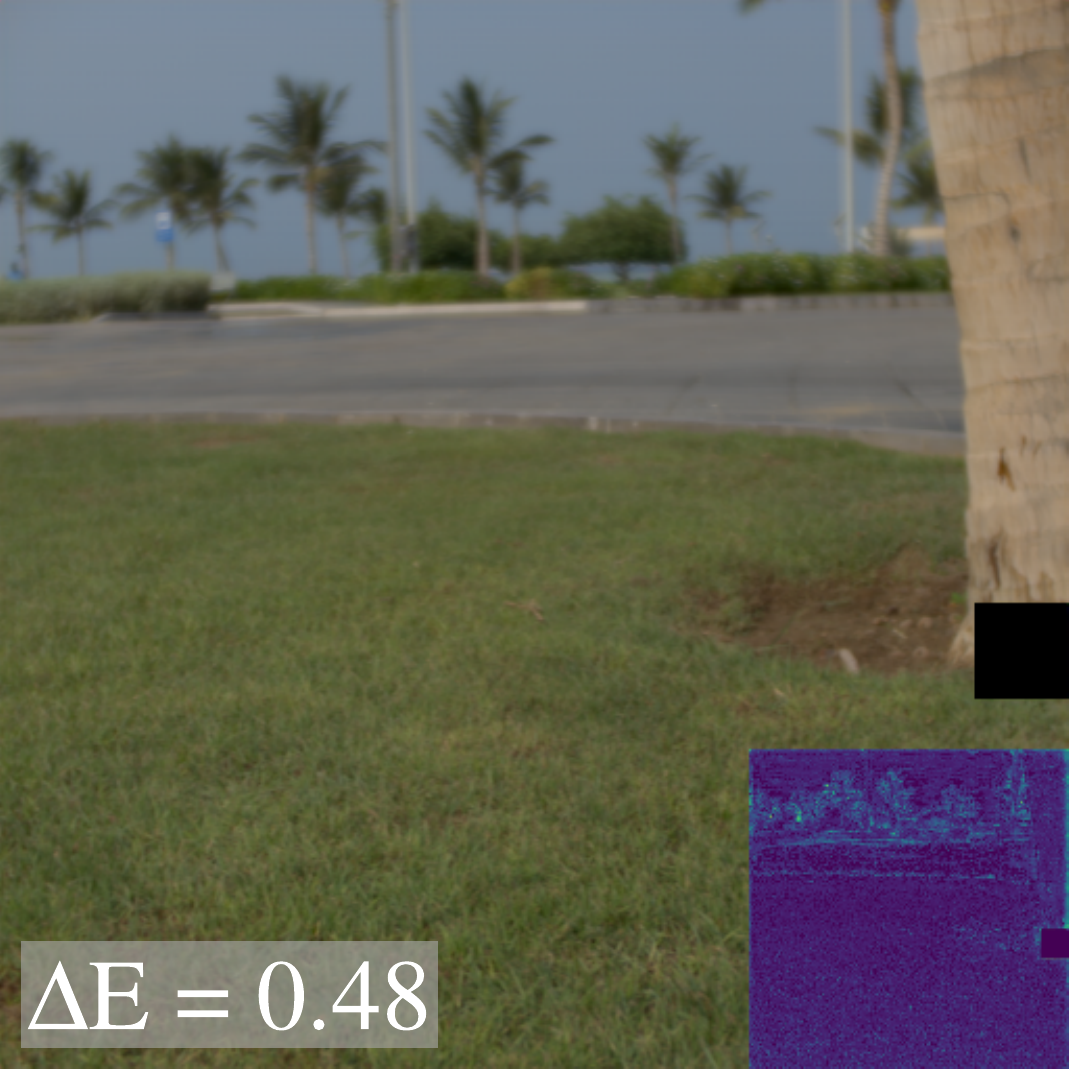} 
    \end{subfigure}
    \begin{subfigure}{0.135\linewidth}
    \includegraphics[width=\linewidth]{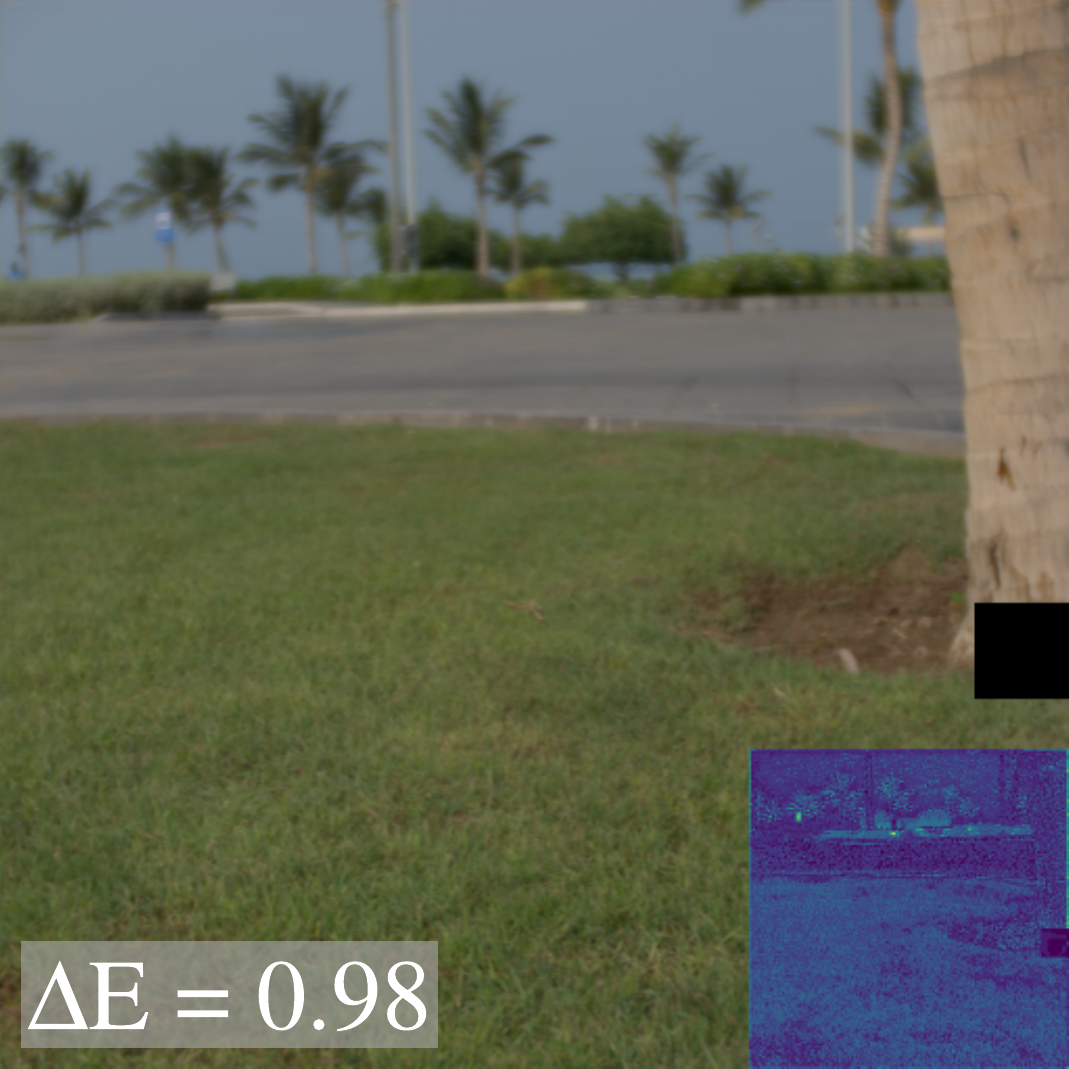}
    \end{subfigure}
    \begin{subfigure}{0.135\linewidth}
    \includegraphics[width=\linewidth]{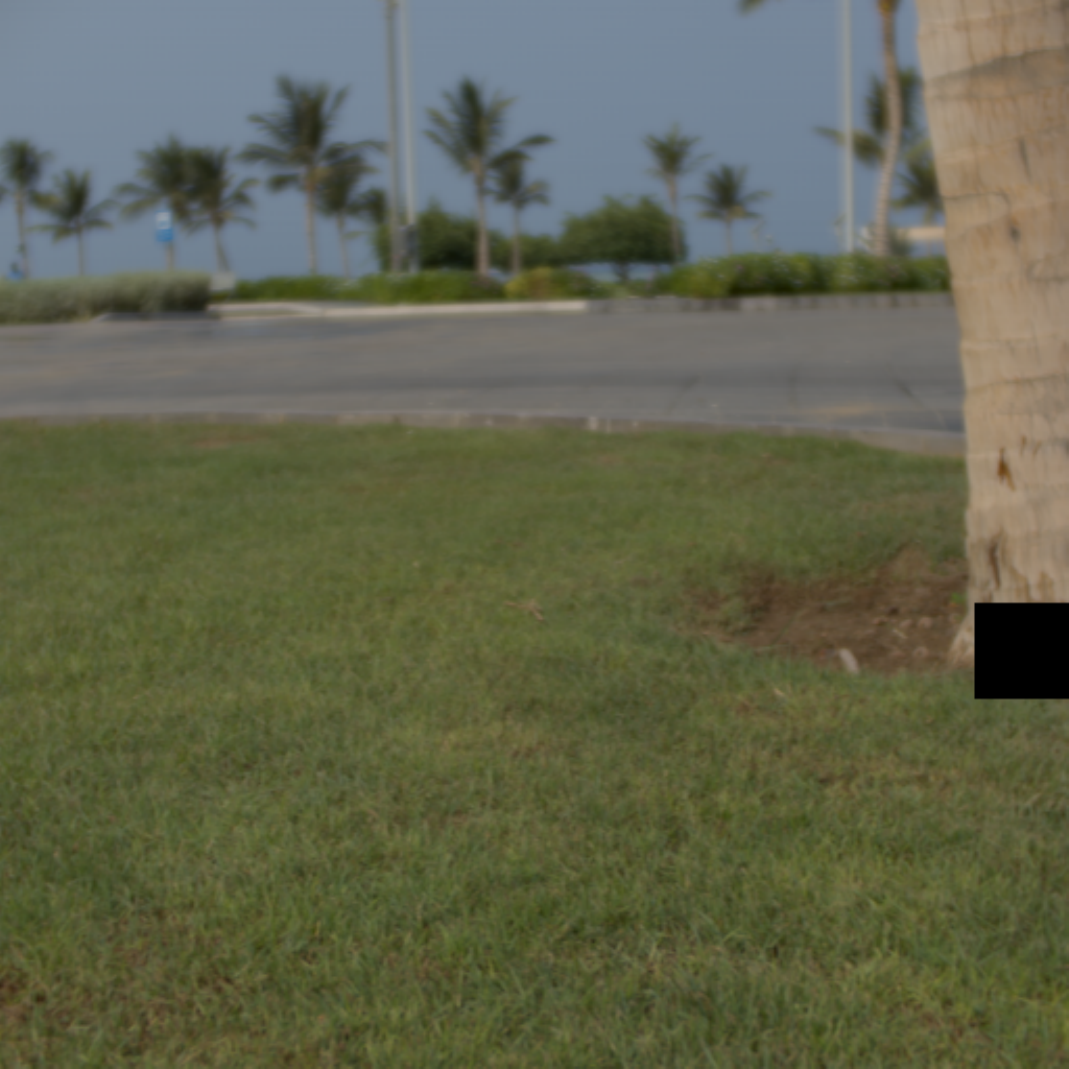} 
    \end{subfigure}

    \begin{subfigure}{0.135\linewidth}
    \includegraphics[width=\linewidth]{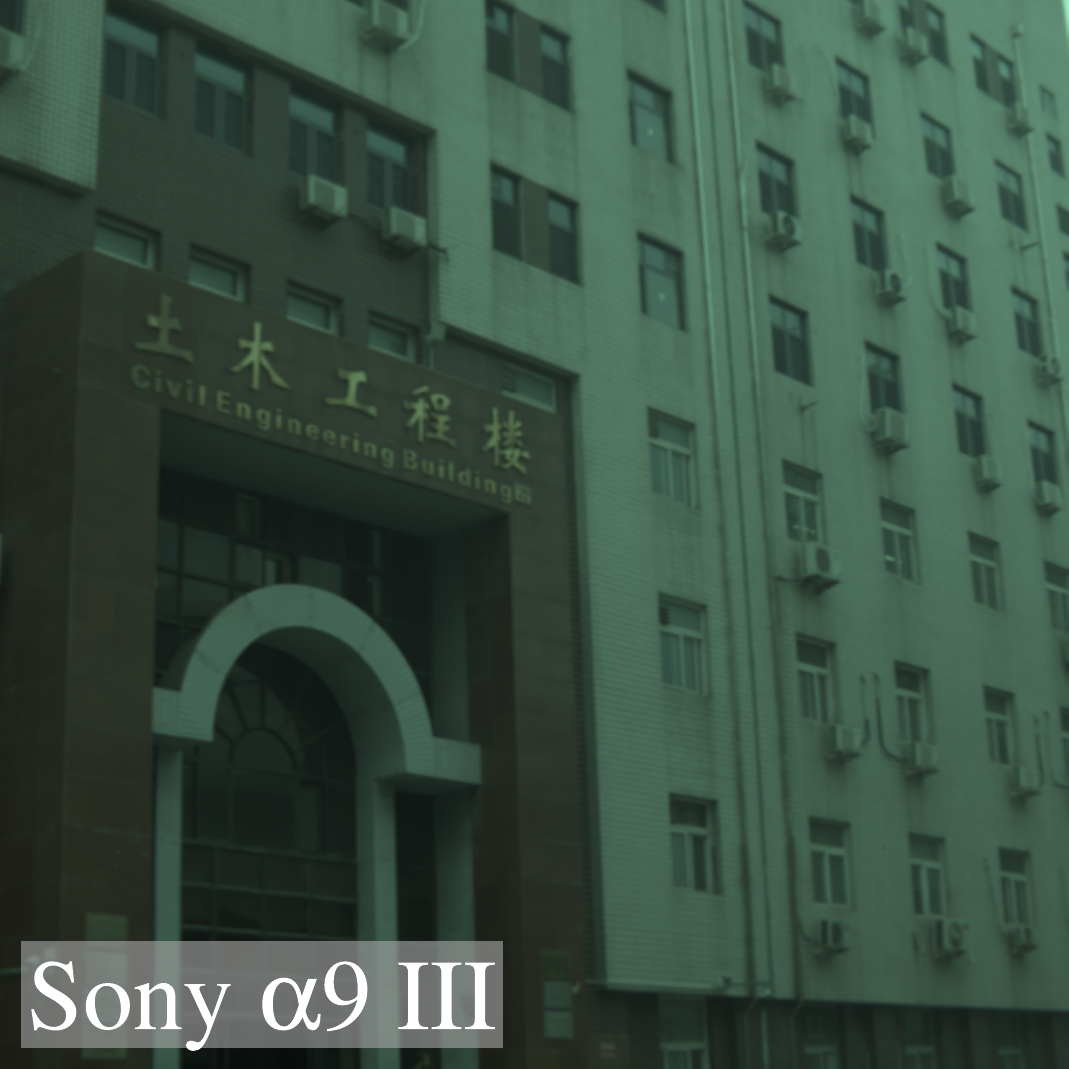} 
    \end{subfigure}
    \begin{subfigure}{0.0078\linewidth}
    \includegraphics[width=\linewidth]{assets/qualitative_main+dE5/cmap_scale.pdf} 
    \end{subfigure}
    \begin{subfigure}{0.135\linewidth}
    \includegraphics[width=\linewidth]{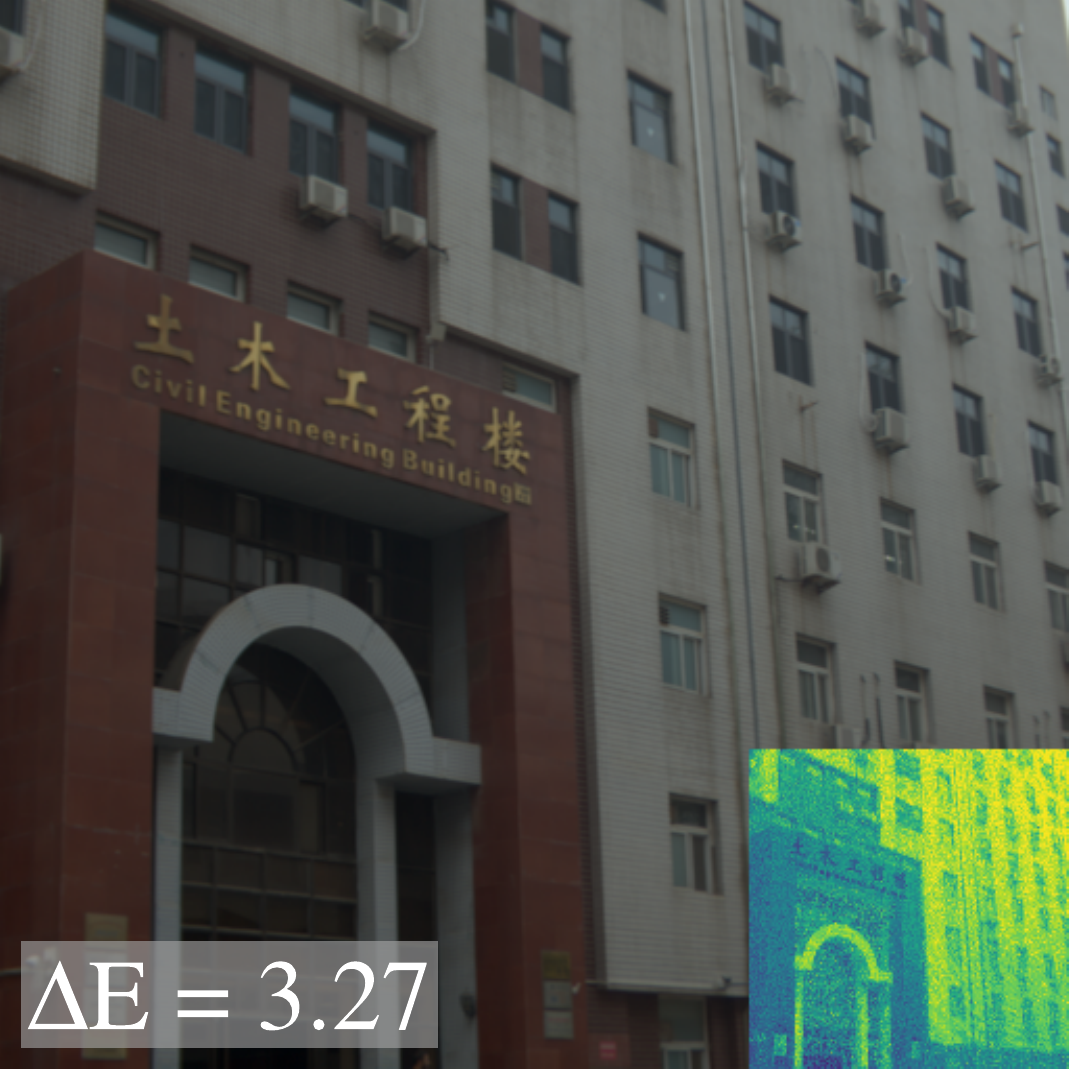} 
    \end{subfigure}
    \begin{subfigure}{0.135\linewidth}
    \includegraphics[width=\linewidth]{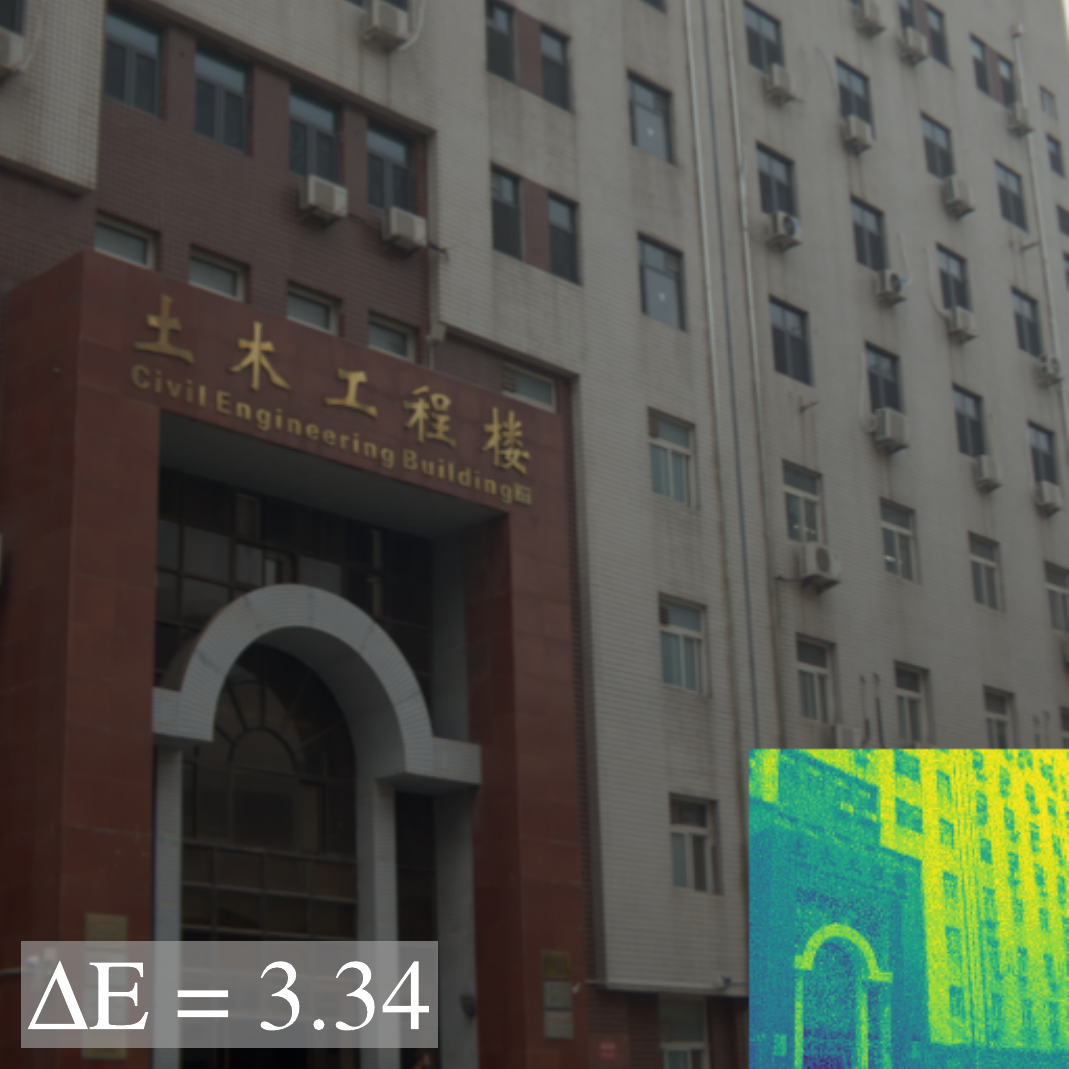} 
    \end{subfigure}
    \begin{subfigure}{0.135\linewidth}
    \includegraphics[width=\linewidth]{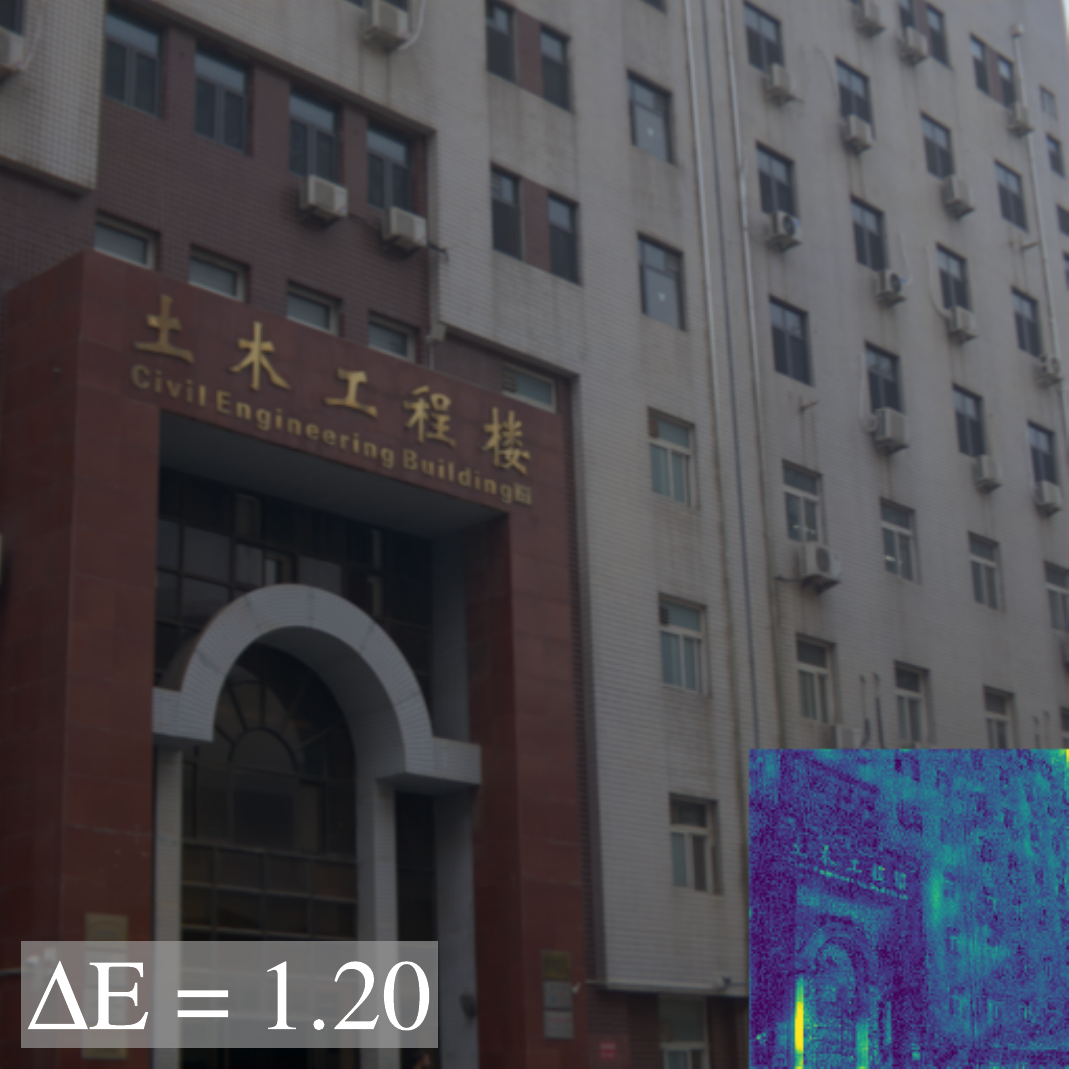} 
    \end{subfigure}
    \begin{subfigure}{0.135\linewidth}
    \includegraphics[width=\linewidth]{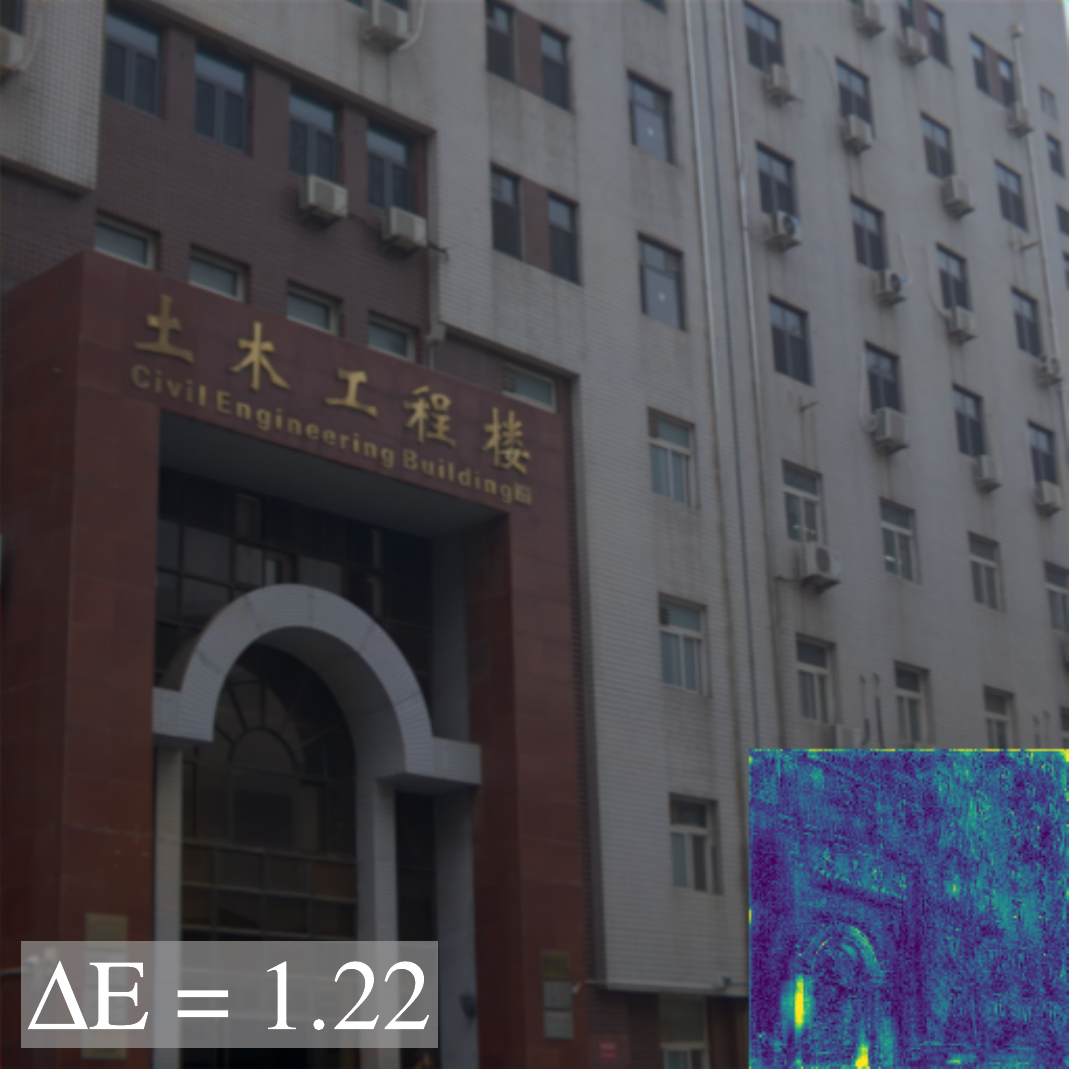} 
    \end{subfigure}
    \begin{subfigure}{0.135\linewidth}
    \includegraphics[width=\linewidth]{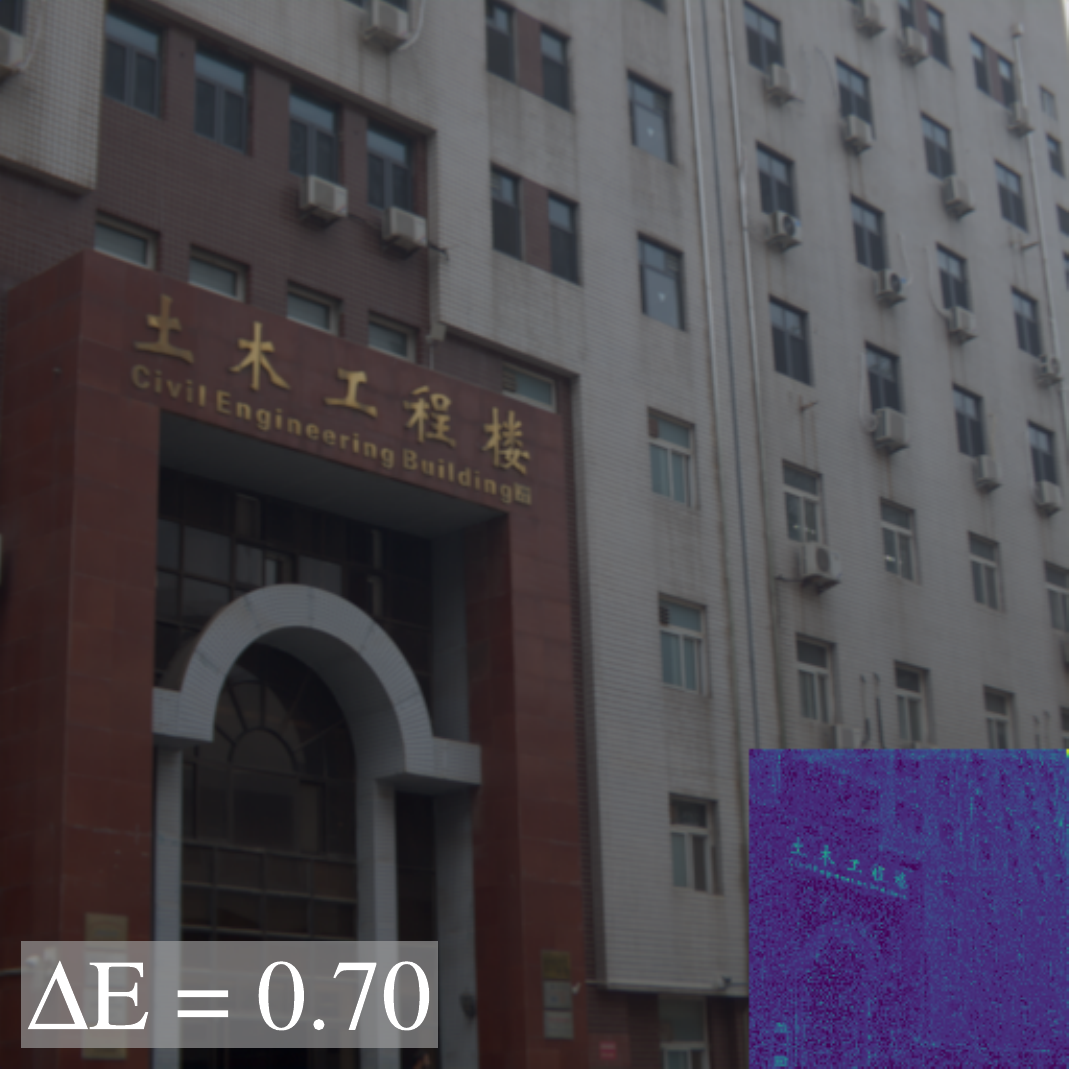}
    \end{subfigure}
    \begin{subfigure}{0.135\linewidth}
    \includegraphics[width=\linewidth]{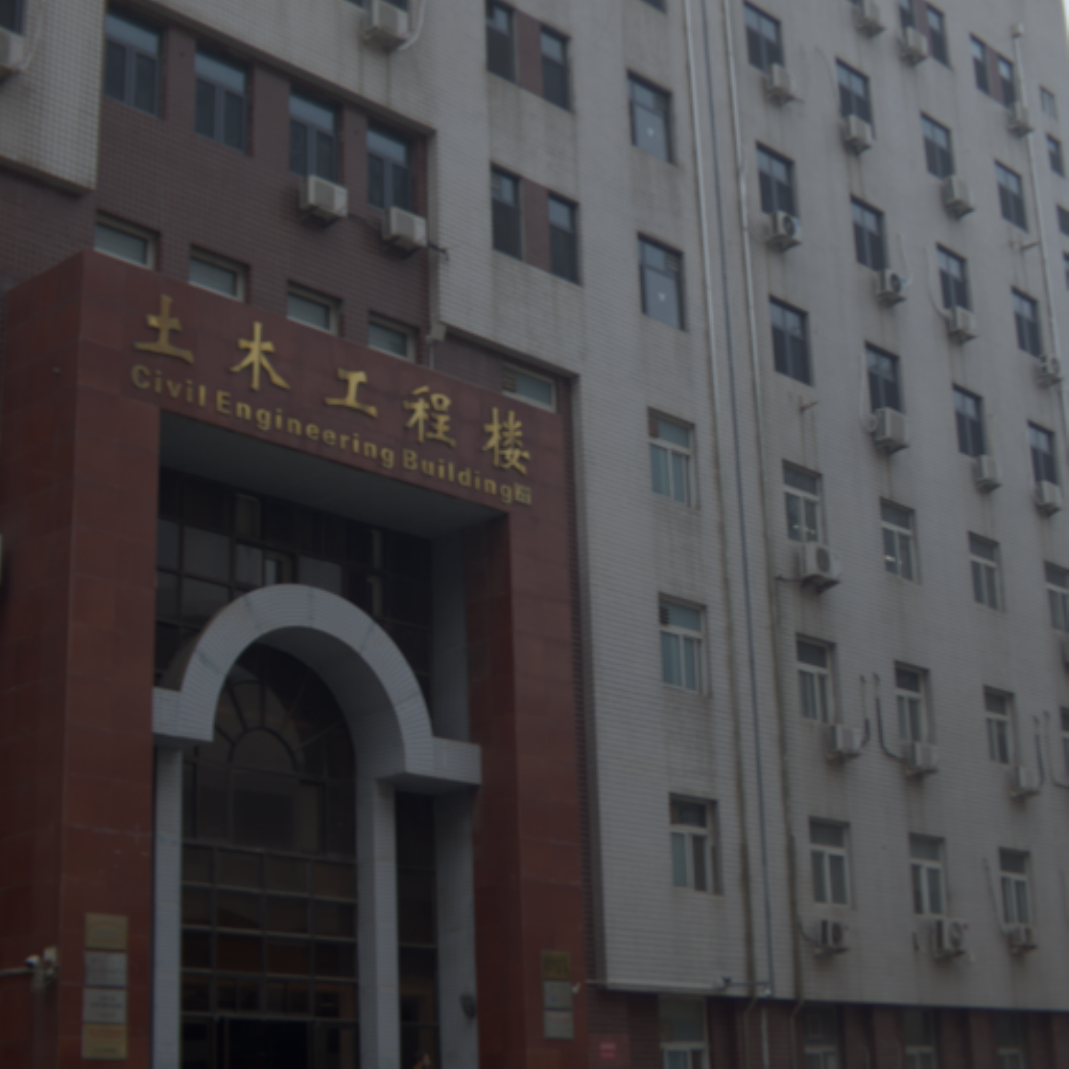} 
    \end{subfigure}

    \begin{subfigure}{0.135\linewidth}
    \includegraphics[width=\linewidth]{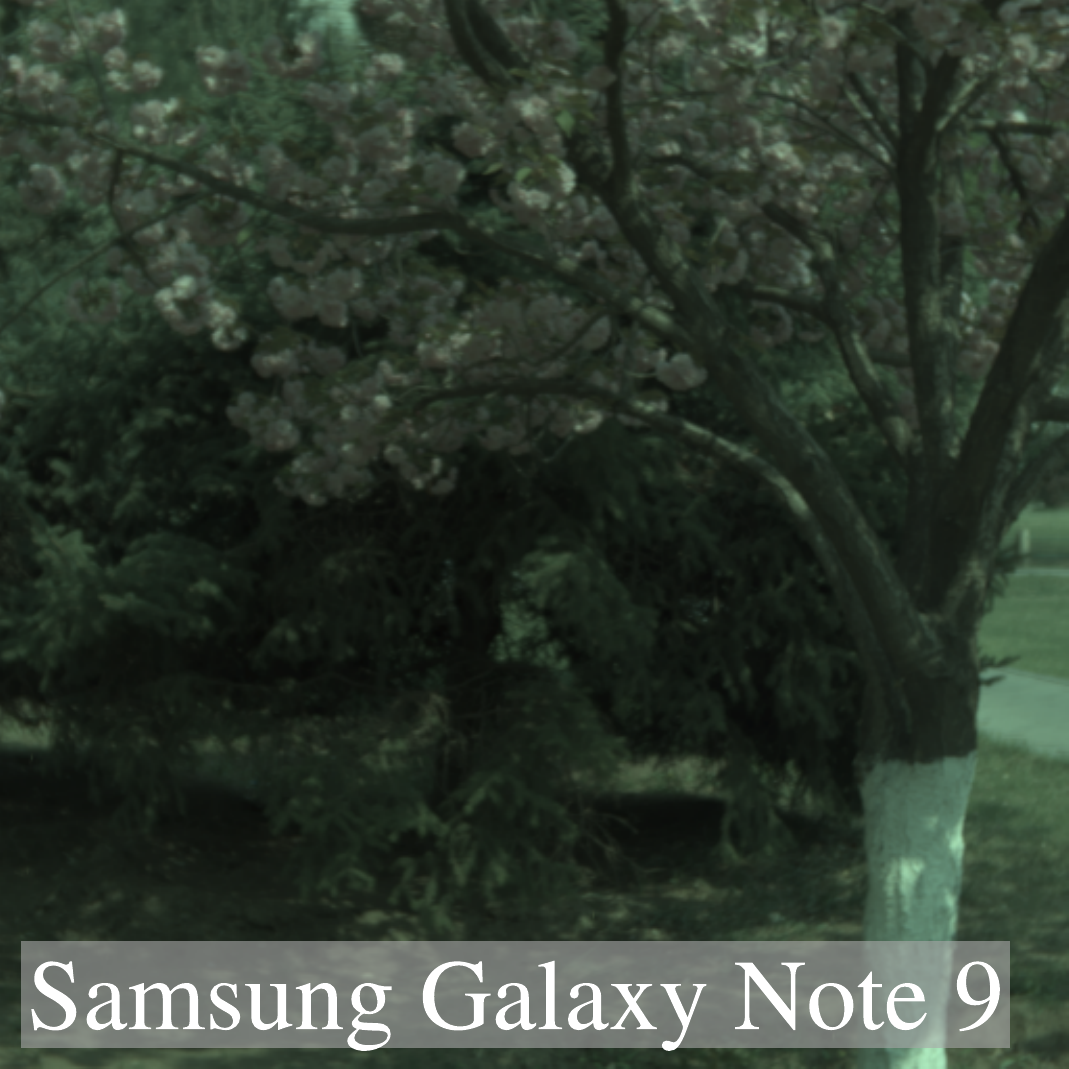} 
    \end{subfigure}
    \begin{subfigure}{0.0078\linewidth}
    \includegraphics[width=\linewidth]{assets/qualitative_main+dE5/cmap_scale.pdf} 
    \end{subfigure}
    \begin{subfigure}{0.135\linewidth}
    \includegraphics[width=\linewidth]{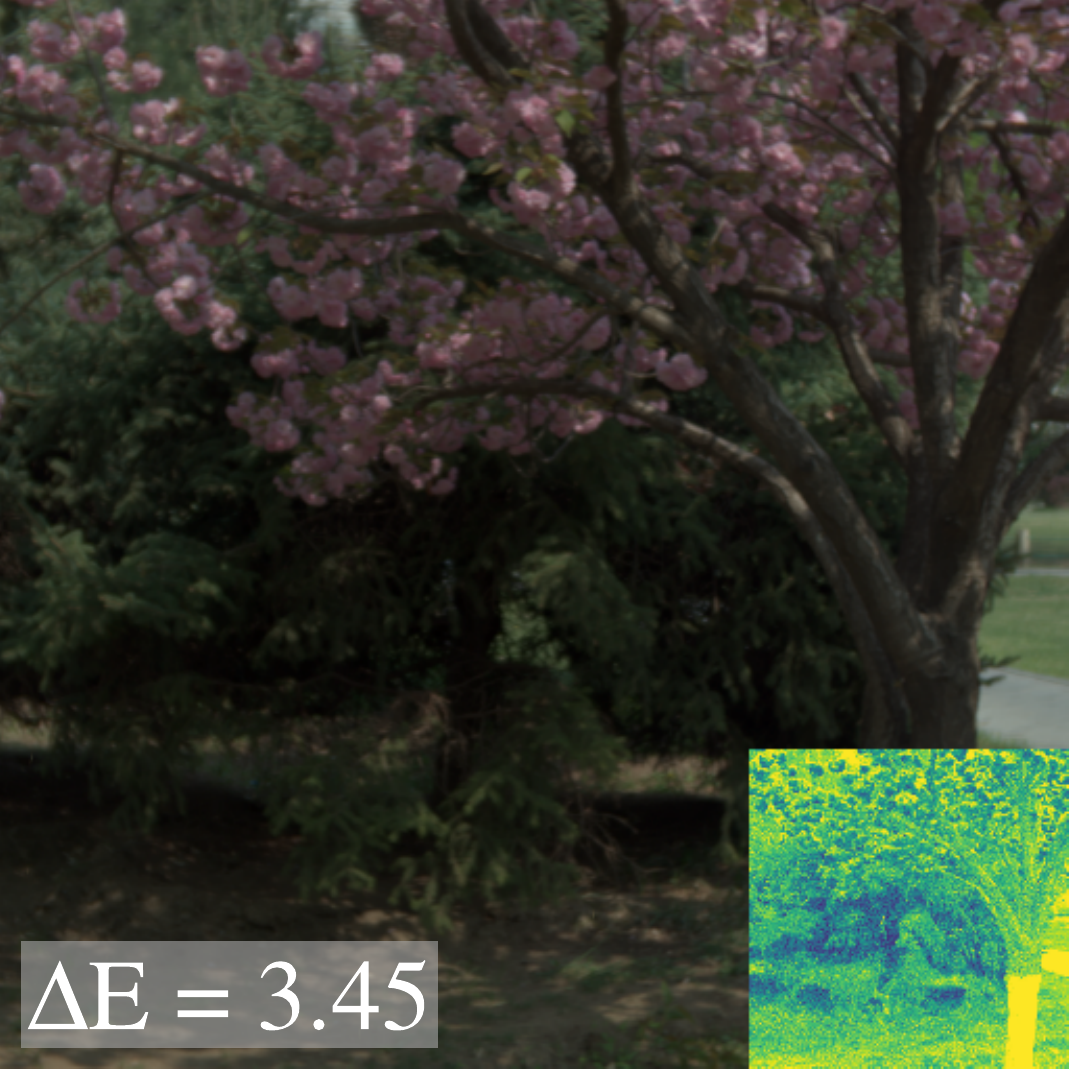} 
    \end{subfigure}
    \begin{subfigure}{0.135\linewidth}
    \includegraphics[width=\linewidth]{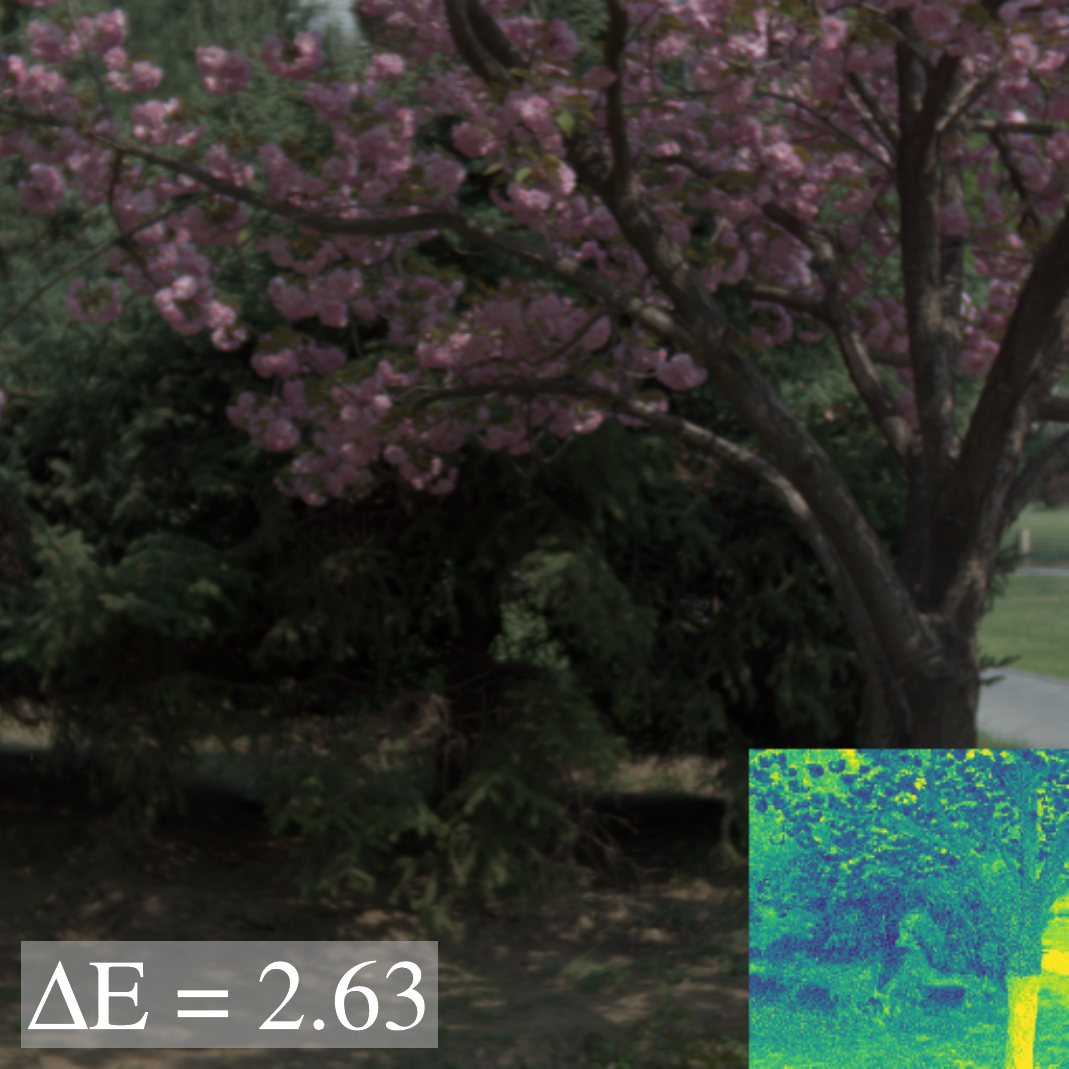} 
    \end{subfigure}
    \begin{subfigure}{0.135\linewidth}
    \includegraphics[width=\linewidth]{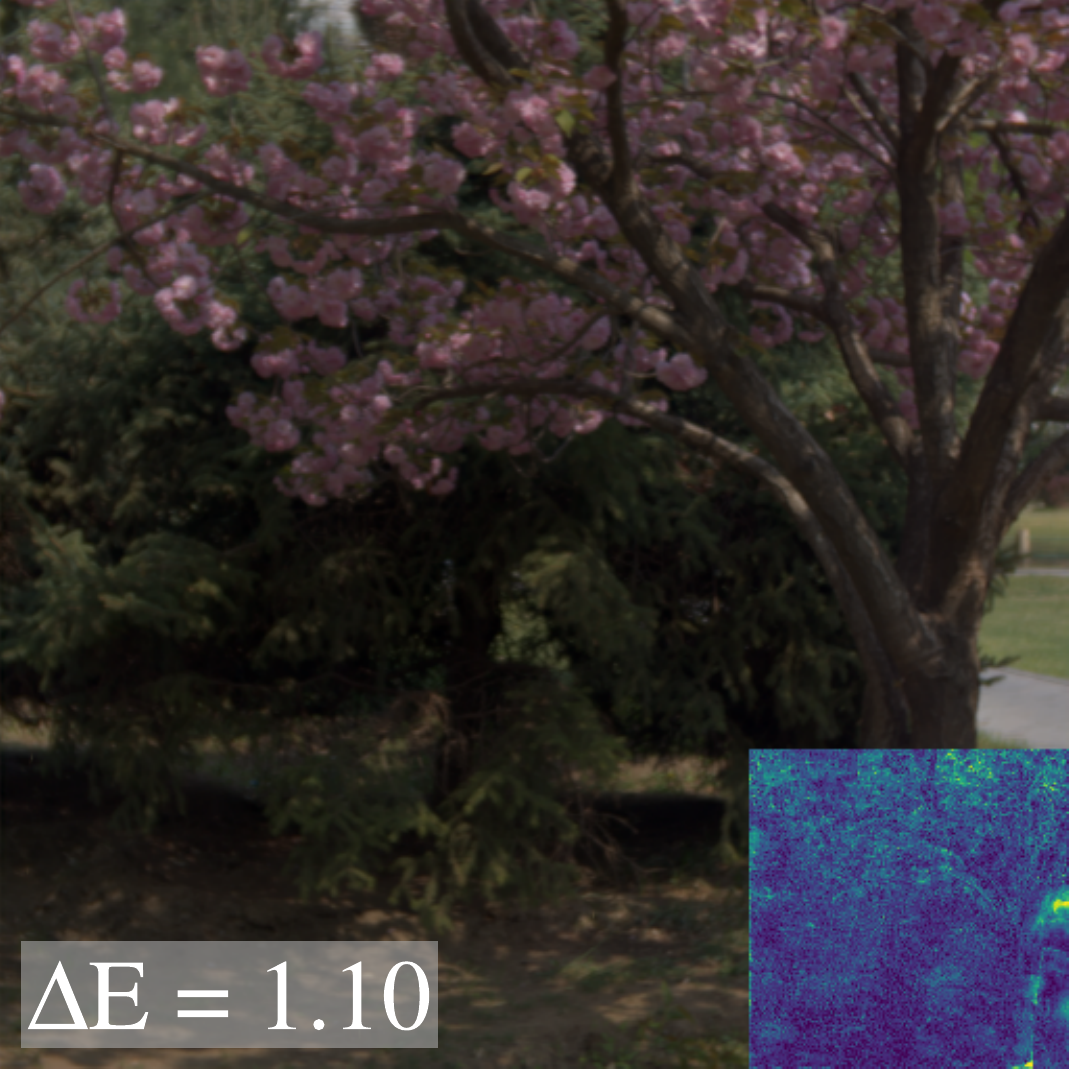} 
    \end{subfigure}
    \begin{subfigure}{0.135\linewidth}
    \includegraphics[width=\linewidth]{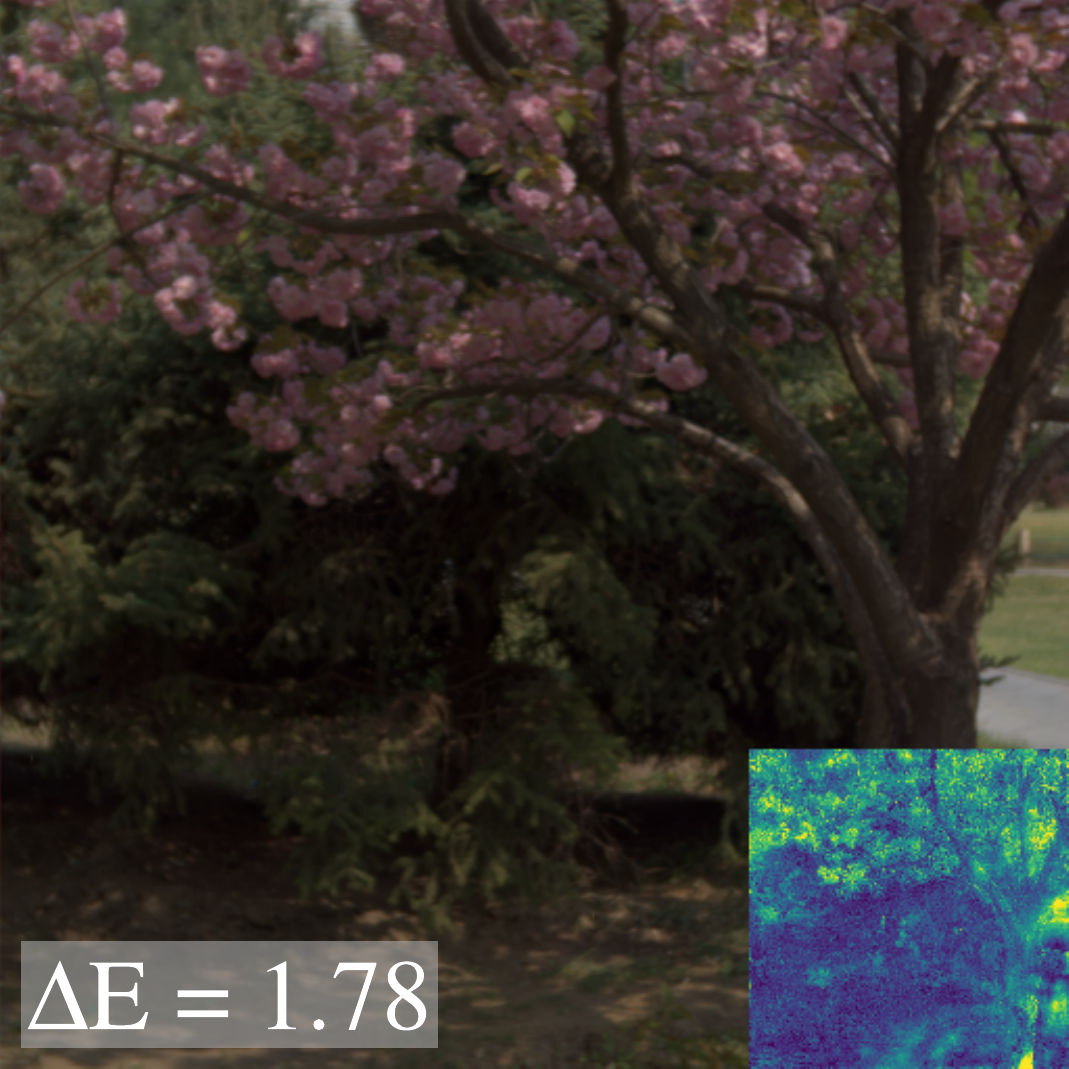} 
    \end{subfigure}
    \begin{subfigure}{0.135\linewidth}
    \includegraphics[width=\linewidth]{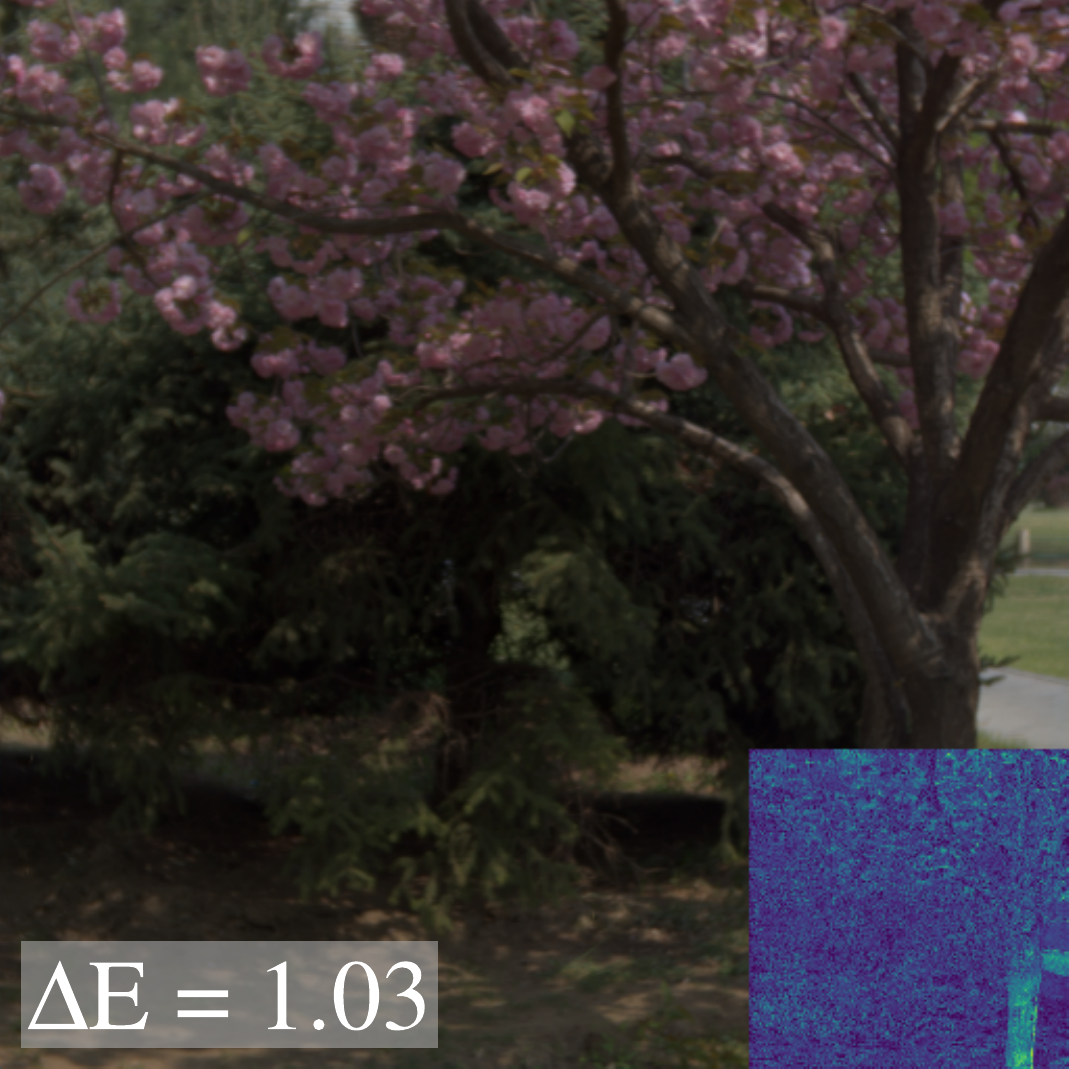}
    \end{subfigure}
    \begin{subfigure}{0.135\linewidth}
    \includegraphics[width=\linewidth]{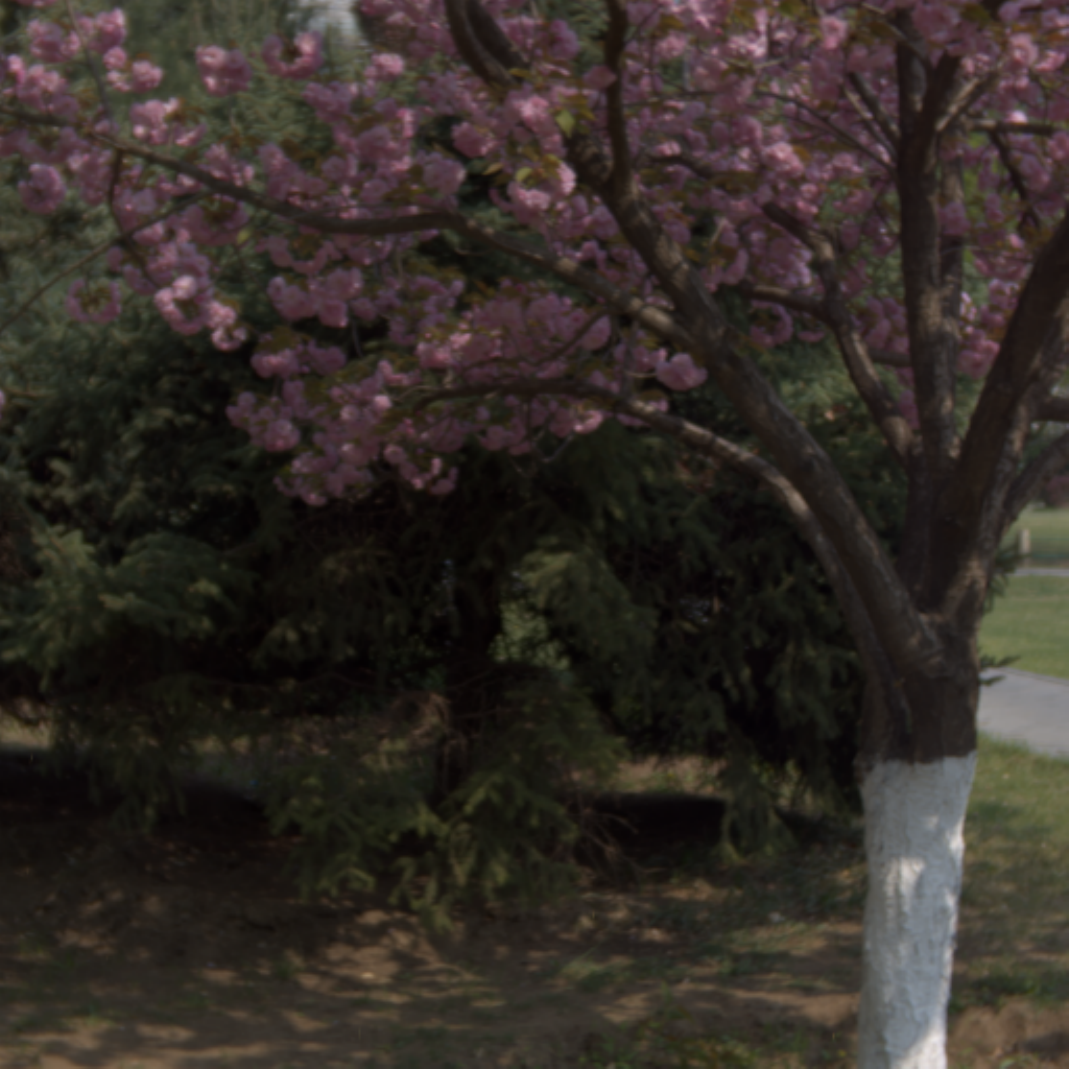} 
    \end{subfigure}

    \begin{subfigure}{0.135\linewidth}
    \includegraphics[width=\linewidth]{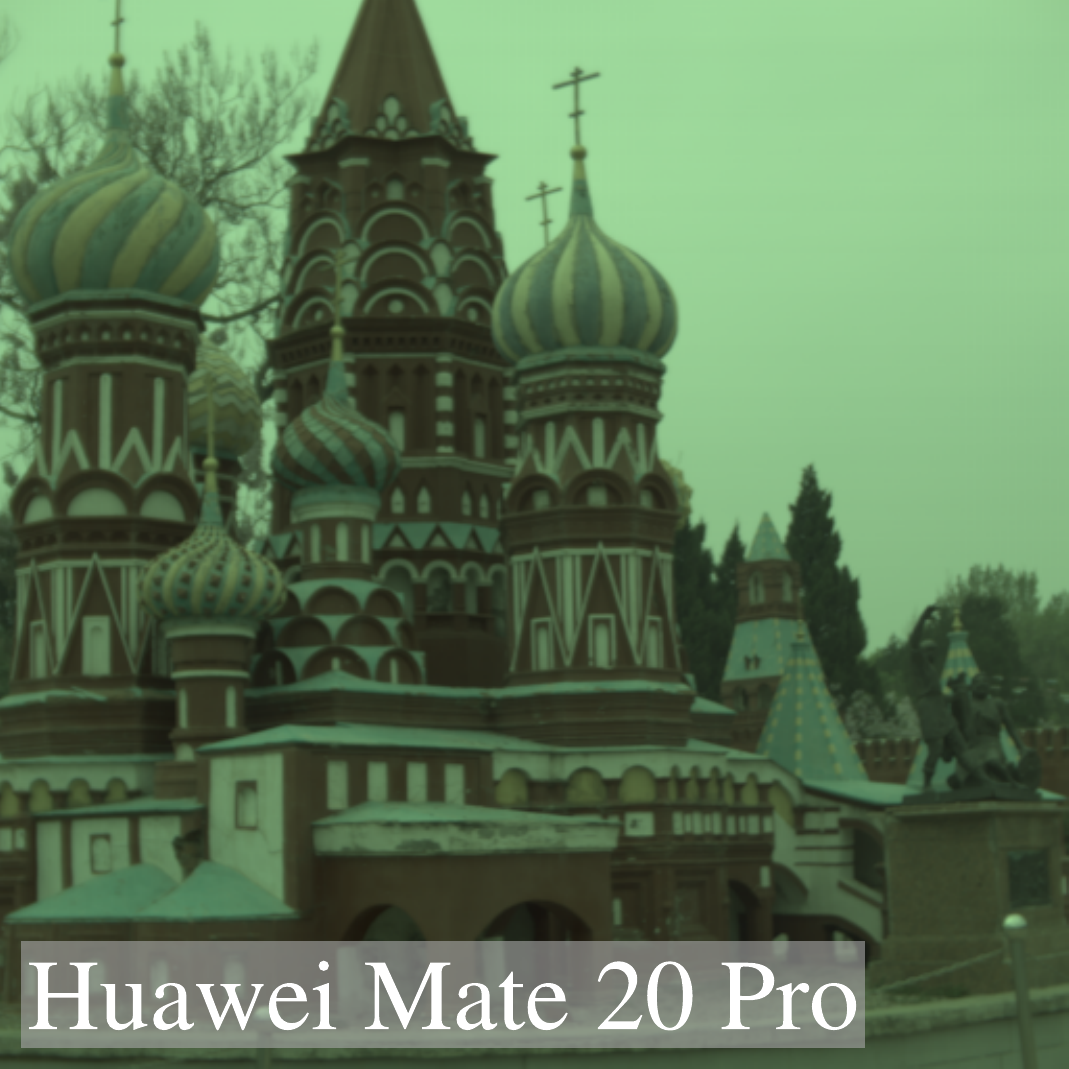} 
    \end{subfigure}
    \begin{subfigure}{0.0078\linewidth}
    \includegraphics[width=\linewidth]{assets/qualitative_main+dE5/cmap_scale.pdf} 
    \end{subfigure}
    \begin{subfigure}{0.135\linewidth}
    \includegraphics[width=\linewidth]{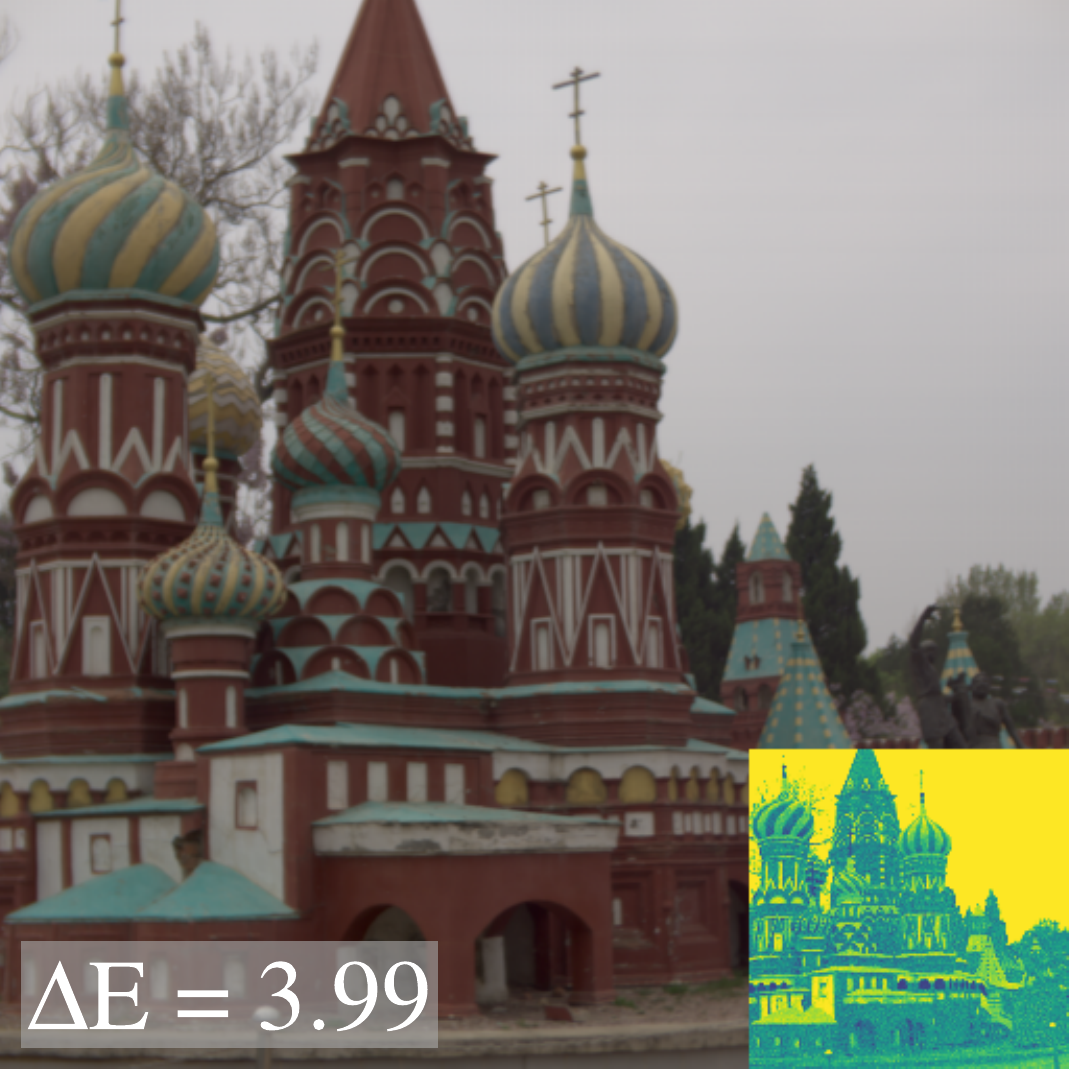} 
    \end{subfigure}
    \begin{subfigure}{0.135\linewidth}
    \includegraphics[width=\linewidth]{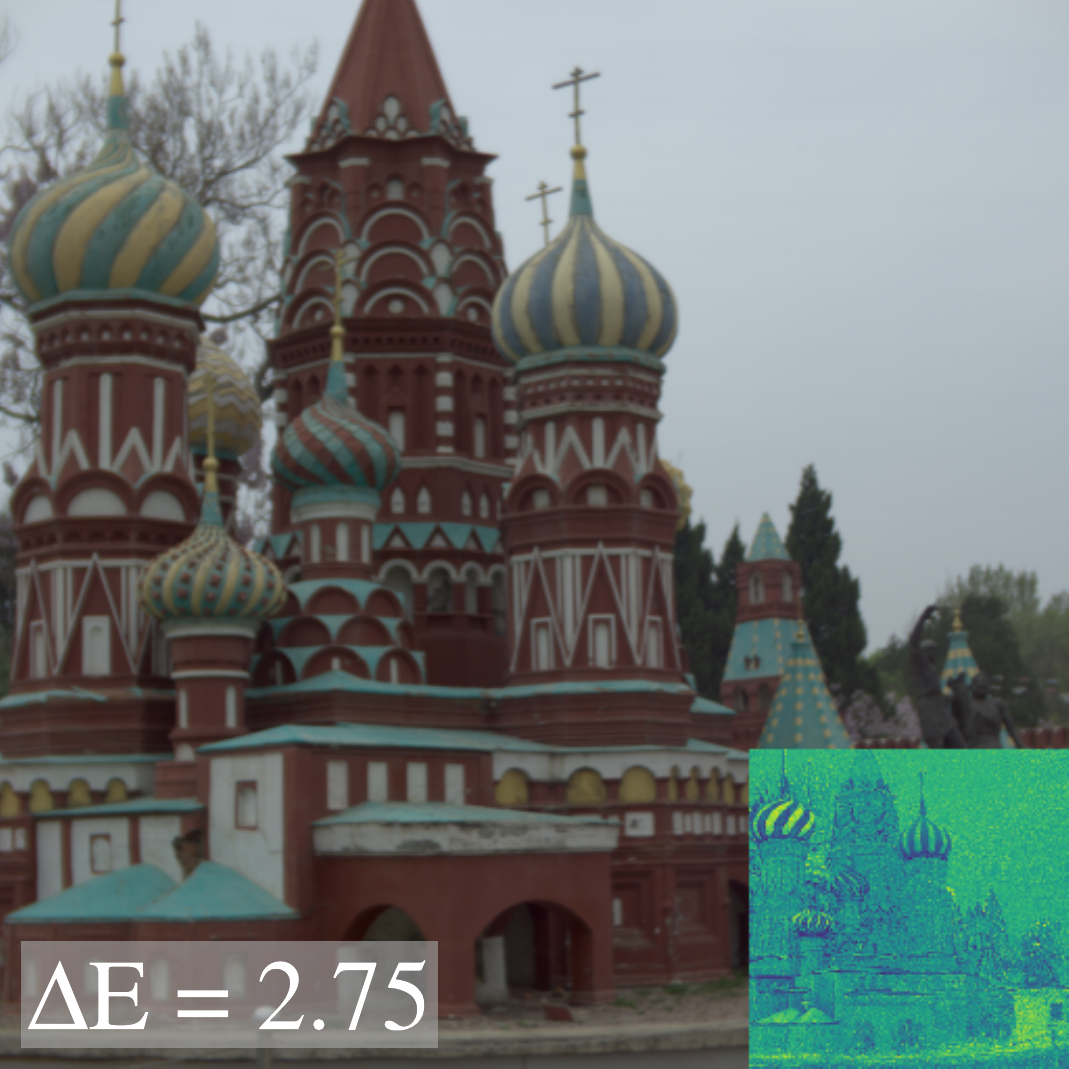} 
    \end{subfigure}
    \begin{subfigure}{0.135\linewidth}
    \includegraphics[width=\linewidth]{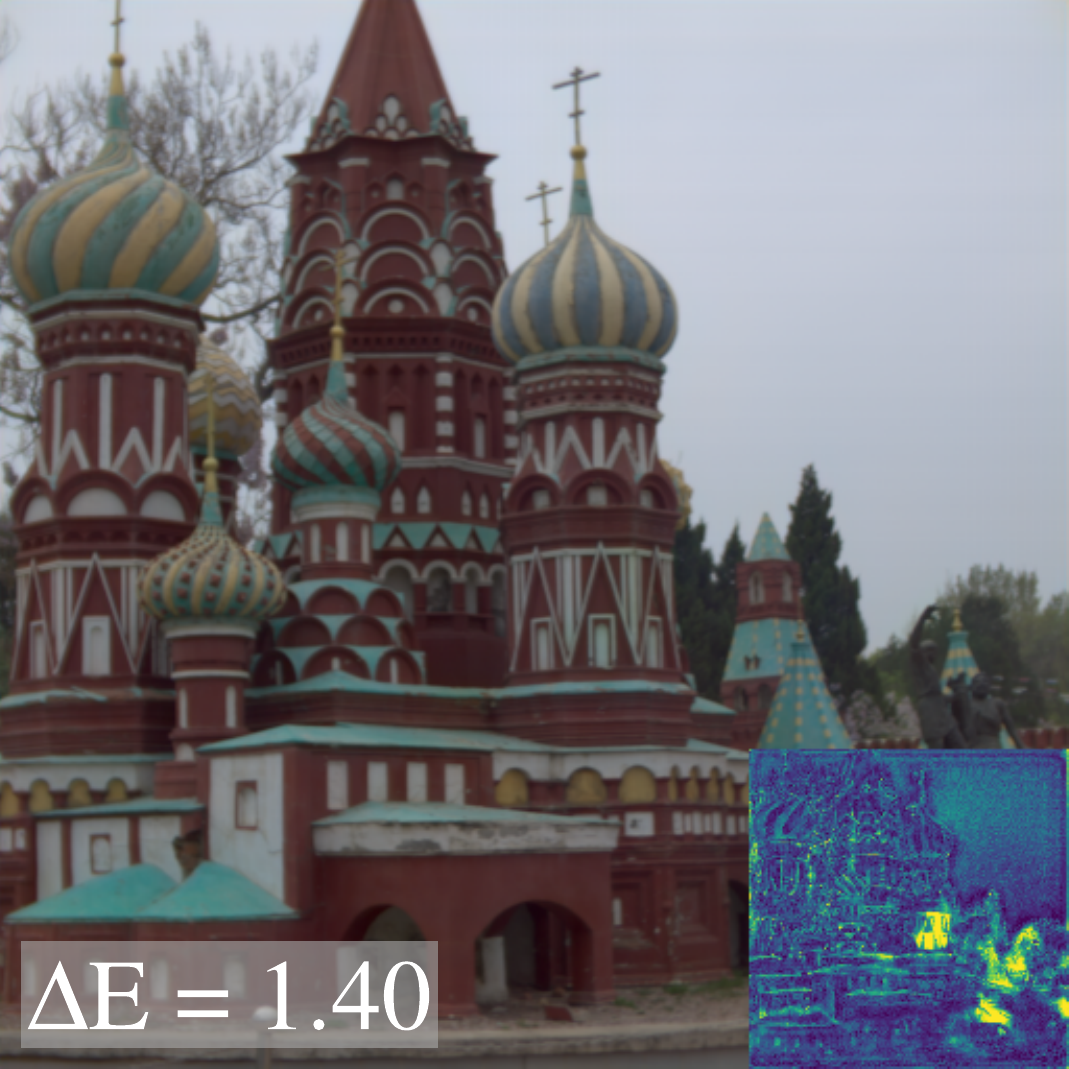} 
    \end{subfigure}
    \begin{subfigure}{0.135\linewidth}
    \includegraphics[width=\linewidth]{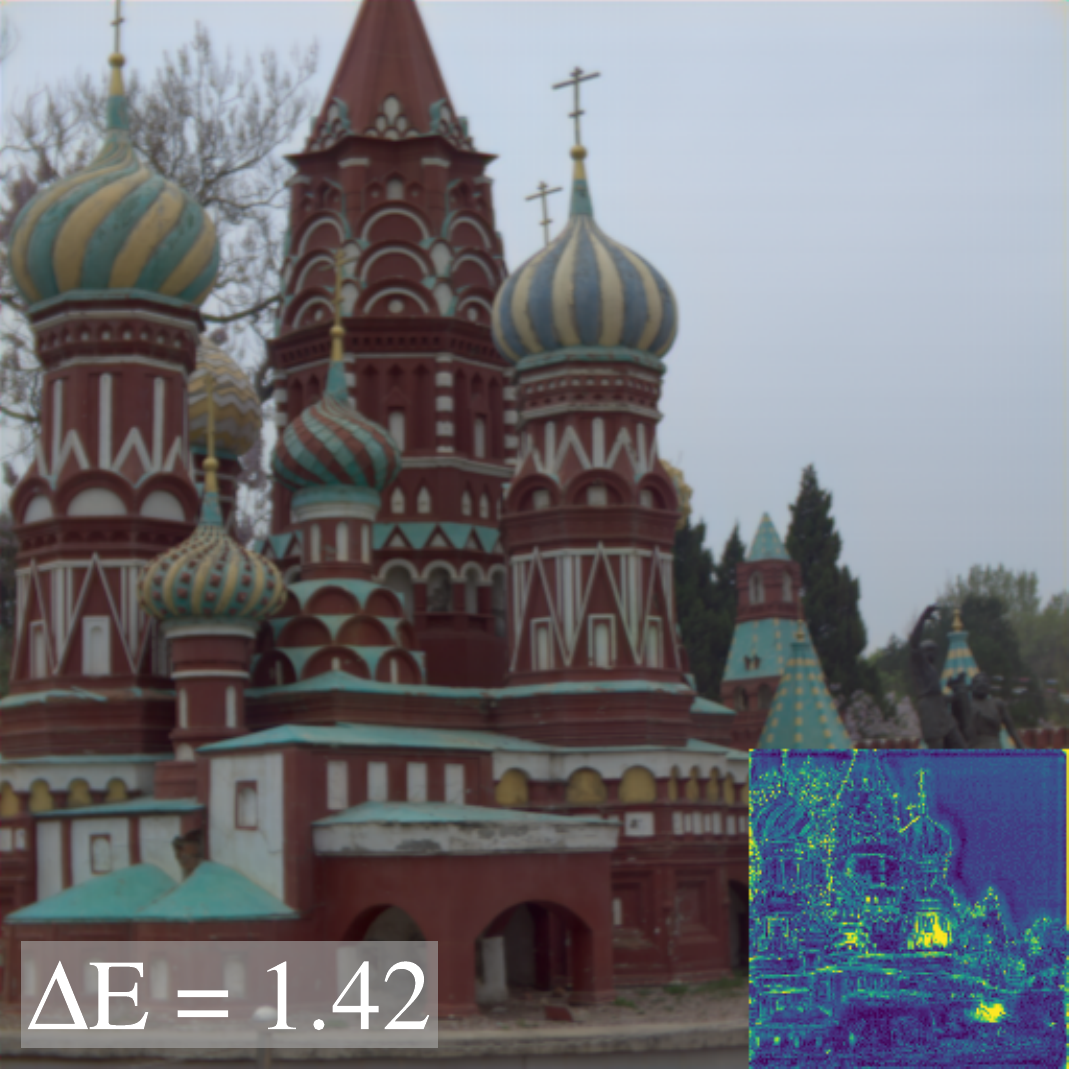} 
    \end{subfigure}
    \begin{subfigure}{0.135\linewidth}
    \includegraphics[width=\linewidth]{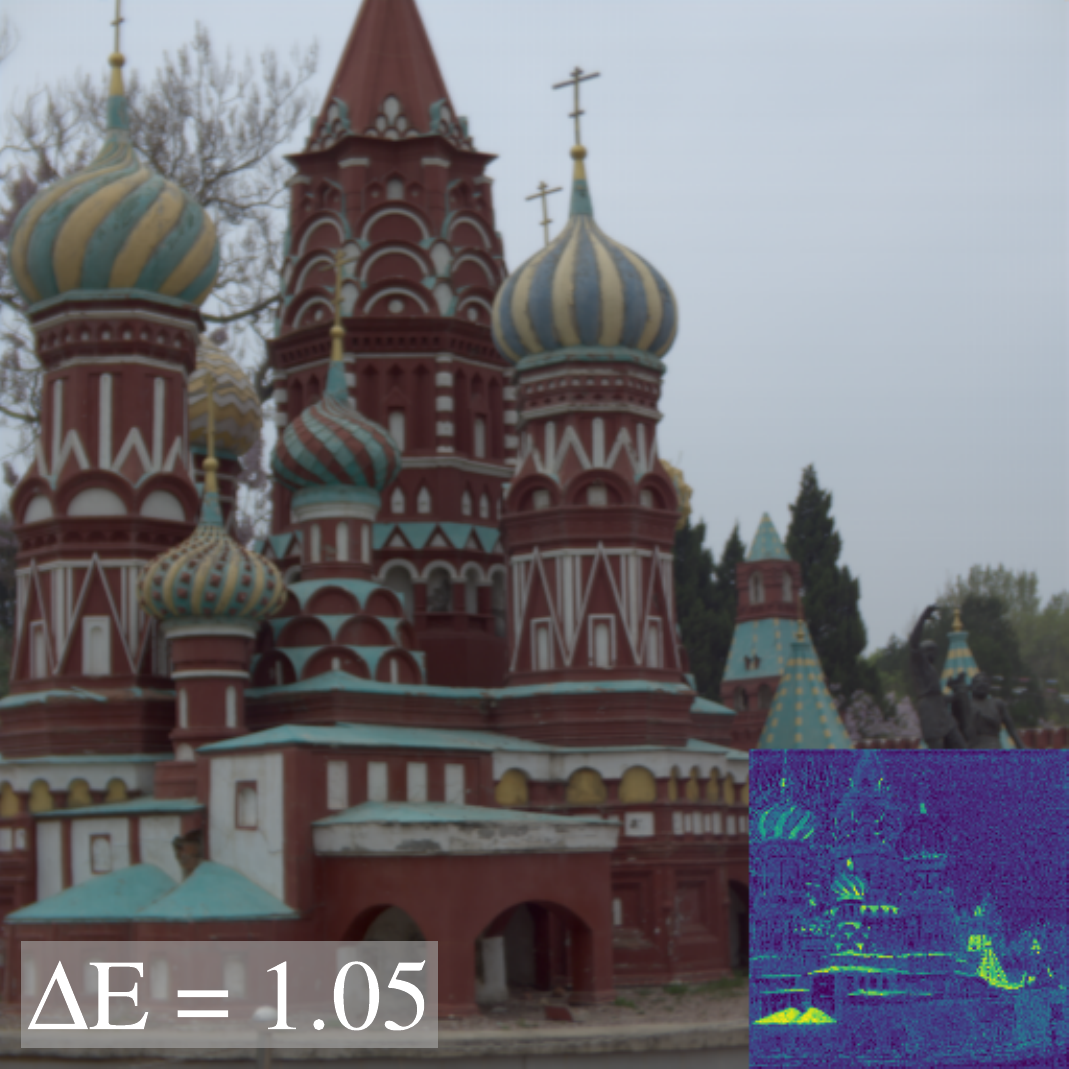}
    \end{subfigure}
    \begin{subfigure}{0.135\linewidth}
    \includegraphics[width=\linewidth]{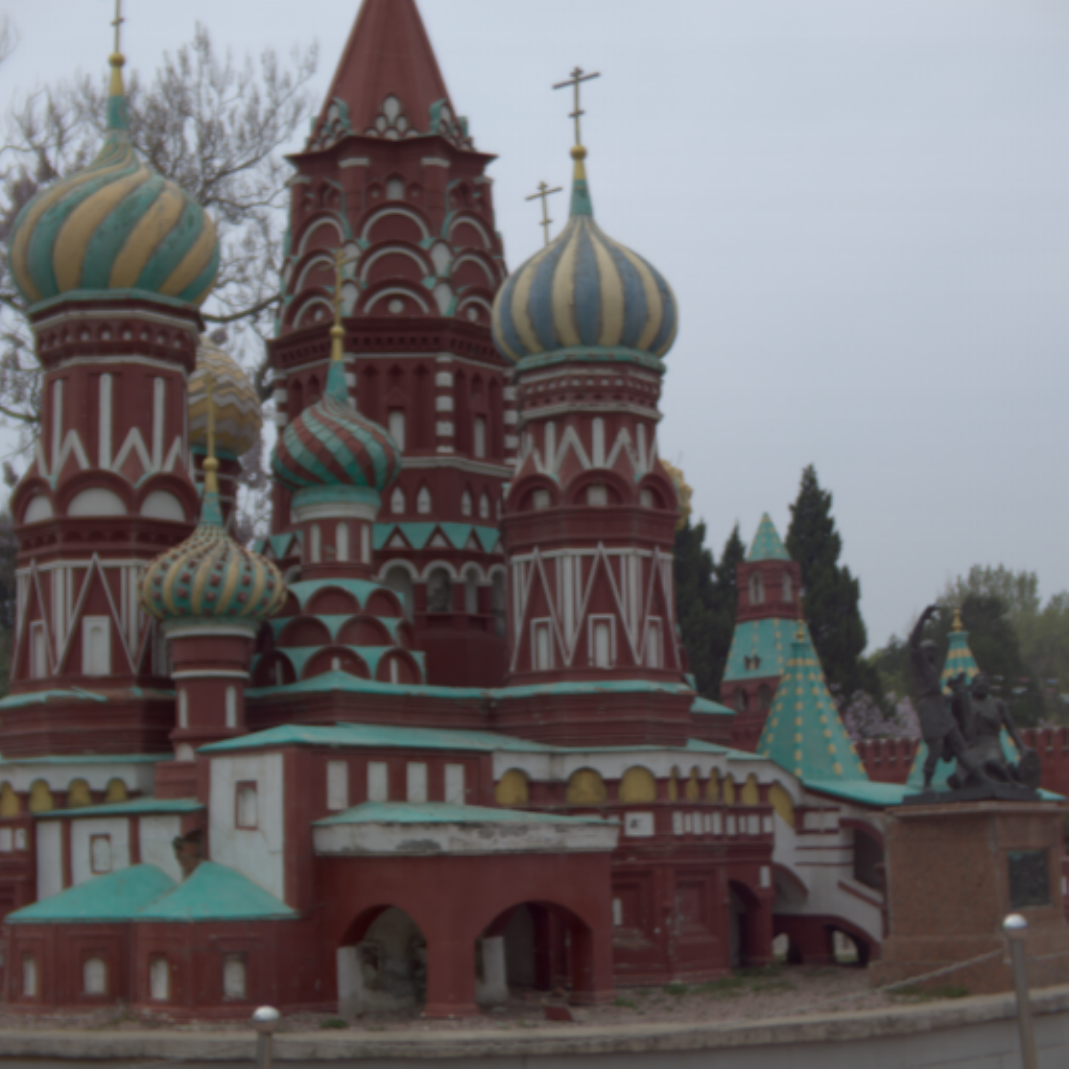} 
    \end{subfigure}

    \vspace{-10pt}
    \subcaptionbox{\scriptsize raw RGB}[0.135\linewidth]{}
    \subcaptionbox{}[0.0078\linewidth]{}
    \subcaptionbox{\scriptsize FC$^4$~\cite{hu2017fc4}}[0.135\linewidth]{}
    \subcaptionbox{\scriptsize SpectralConvMean~\cite{Gong2019ConvolutionalMA}}[0.135\linewidth]{}
    \subcaptionbox{\scriptsize LPIENet (Ours)}[0.135\linewidth]{}
    \subcaptionbox{\scriptsize LPIENet-small (Ours)}[0.135\linewidth]{}
    \subcaptionbox{\scriptsize cmKAN-light (Ours)}[0.135\linewidth]{}
    \subcaptionbox{\scriptsize GT}[0.135\linewidth]{}
      
\caption{Qualitative results of the best performing methods: FC$^{4}$~\cite{hu2017fc4}, SpectralConvMean~\cite{Gong2019ConvolutionalMA}, and our three proposed models. We show results from two mirrorless and two mobile cameras, with $\Delta$E$_{00}$ map reported in the bottom-right corner and average $\Delta$E$_{00}$ reported in the bottom-left one. For visualization purposes, we gamma correct the raw visualizations (first column) and convert the results (columns two to six) to sRGB. More qualitative results are reported in the Supplementary Material.}
    \label{fig:main_qualitative}
\end{figure*}

\begin{table*}[ht]

\caption{Results on the misaligned version of the proposed dataset. Aggregation by camera type, metric reporting, and highlighting follow the same format as in \Cref{tab:main_results}. Our proposed models are adapted to the new input by fine-tuning their spectral encoder, while MS-driven baselines are completely retrained. RGB-only methods are omitted as they are unaffected by spatial misregistration. Per-camera results are provided in the Supplementary Material.}
\label{tab:misaligned_results}
\resizebox{\linewidth}{!}{%
\begin{tabular}{llcccccccccccccccc}
    \toprule
                                            &                                           & \multicolumn{8}{c}{$\Delta$E$_{00}$ $\downarrow$}   & \multicolumn{8}{c}{Reproduction Error $\downarrow$} \\                  \cmidrule(lr){3-10} \cmidrule(lr){11-18} \\\\\noalign{\vskip -2.25em}
    Camera                                  & Method                                    & Mean                  & Med.             & Tri.                & B-25              & W-25                  & 95-P                  & 99-P                  & Max          & Mean                  & Med.             & Tri.                & B-25              & W-25                  & 95-P                  & 99-P                  & Max               \\ \midrule
    \multirow{5}{*}{\vtop{\hbox{\strut Mirrorless}\hbox{\strut Sensors}}} 
        & SpectralFC$^4$~\cite{hu2017fc4}                 & 3.25                      & 2.79                      & 2.88                      & 1.35                      & 5.93                      & 6.91                      & 10.47                     & 23.17                      & 4.34                      & 2.73                      & 3.13                      & 0.90                      & 10.47                     & 13.53                      & 18.61                      & 39.57                      \\
& SpectralConvMean~\cite{Gong2019ConvolutionalMA} & 3.18                      & 2.78                      & 2.86                      &  \tbest{1.34}                & 5.72                      & 6.61                      & 10.00                     &  \tbest{20.13}                & 4.26                      & 2.71                      & 3.05                      &  \tbest{0.89}                & 10.30                     & 13.21                      & 18.24                      & 40.76                      \\ \\\noalign{\vskip -1em} \cdashline{2-18} \\\noalign{\vskip -1em}
& LPIENet~(Ours)              & \best{1.84}             & \sbest{1.65}             & \sbest{1.69}             & \sbest{1.02}             & \best{2.99}             & \best{3.46}             & \best{4.66}             & \best{13.41}             & \sbest{3.39}             & \sbest{1.74}             & \sbest{2.22}             & \sbest{0.74}             & \sbest{8.92}             & \best{10.81}             & \sbest{17.04}             & \best{38.94}             \\
& LPIENet-small~(Ours)        & \sbest{2.34}             &  \tbest{2.14}                &  \tbest{2.17}                & 1.40                      &  \tbest{3.66}                &  \tbest{4.17}                & \sbest{5.78}             & \sbest{19.95}             &  \tbest{3.86}                &  \tbest{2.21}                &  \tbest{2.67}                & 1.02                      &  \tbest{9.56}                &  \tbest{11.73}                &  \tbest{17.44}                & \sbest{39.42}             \\
& cmKAN-light~(Ours)           & \best{1.84}             & \best{1.51}             & \best{1.58}             & \best{0.82}             & \sbest{3.44}             & \sbest{4.07}             &  \tbest{6.26}                & 20.93                      & \best{3.20}             & \best{1.50}             & \best{1.96}             & \best{0.50}             & \best{8.75}             & \sbest{11.10}             & \best{16.99}             &  \tbest{39.55}                \\ \midrule
    \multirow{5}{*}{\vtop{\hbox{\strut Mobile}\hbox{\strut Sensors}}} 
        & SpectralFC$^4$~\cite{hu2017fc4}                   & 3.23              & 2.74              & 2.83              & 1.43          & 5.88              & 6.92              & 10.50                 & 22.05                 & 4.54              & 2.97             & 3.31             & 1.01             & 10.77                & 13.79                & 19.71                & 39.29 \\
        & SpectralConvMean~\cite{Gong2019ConvolutionalMA}   & 3.12              & 2.69              & 2.78              & 1.37          & 5.59              & 6.45              & 9.95                  & \tbest{20.87}         & 4.31              & 2.82             & 3.15             & \tbest{0.97}             & 10.21                & 12.94                & 18.33                & 40.69 \\ \\\noalign{\vskip -1em} \cdashline{2-18} \\\noalign{\vskip -1em}
        & LPIENet~(Ours)                & \sbest{1.83}      & \sbest{1.62}      & \sbest{1.67}      & \sbest{1.00}  & \best{3.01}       & \best{3.48}       & \best{4.85}           & \best{13.06}          & \sbest{3.36}      & \sbest{1.70}     & \sbest{2.18}     & \sbest{0.75}     & \sbest{8.85}         & \best{10.74}         & \sbest{17.01}        & \sbest{38.41} \\
        & LPIENet-small~(Ours)          & \tbest{2.25}      & \tbest{2.03}      & \tbest{2.07}      & \tbest{1.31}  & \tbest{3.58}      & \tbest{4.10}      & \sbest{5.81}          & \sbest{20.45}         & \tbest{3.77}      & \tbest{2.12}     & \tbest{2.58}     & 0.99     & \tbest{9.39}         & \tbest{11.60}        & \tbest{17.39}        & \sbest{38.41} \\
        & cmKAN-light~(Ours)             & \best{1.75}       & \best{1.43}       & \best{1.50}       & \best{0.78}   & \sbest{3.29}      & \sbest{3.88}      & \tbest{6.31}          & 21.45                 & \best{3.14}       & \best{1.41}      & \best{1.89}      & \best{0.50}      & \best{8.67}          & \sbest{11.14}        & \best{16.88}         & \best{37.89} \\ \bottomrule
    
\end{tabular}
}

\end{table*}

\section{Experiments}
\label{sec:experiments}
\paragraph{Experimental Setting}
\label{subsec:experimental_setting}
All models are trained using the proposed dataset, conducting separate experiments for each simulated RGB sensor. Training is performed with the Adam optimizer~\cite{kingma2014adam}, using a learning rate of $1\times10^{-4}$. We employ early stopping with a maximum of 300 epochs and a patience value of 5, retaining the model that achieves the best performance on the validation set. Cosine annealing scheduling~\cite{loshchilov2016sgdr} is used to adapt the learning rate during training.
Illuminant estimation networks are optimized using the Angular Error as the loss function, while image-to-image models are trained using the $\Delta E_{76}$ color difference metric~\cite{robertson1977cie}. For quantitative evaluation, we compute the $\Delta$E$_{00}$ color distance~\cite{luo2001development} and the Reproduction Error~\cite{finlayson2016reproduction}.

\paragraph{Experiments on Aligned Data}
We compare the proposed framework against several state-of-the-art illuminant estimation methods, including both statistical and learning-based approaches. Statistical baselines are drawn from the edge-based color constancy framework~\cite{van2005color}, comprising Gray-World (GW), White-Patch (WP), Shades-of-Gray (SoG), General Gray-World (GGW), and the first and second-order Gray-Edge variants (GE1 and GE2). Learning-based baselines include FC$^4$~\cite{hu2017fc4}, ConvMean~\cite{Gong2019ConvolutionalMA}, and QU~\cite{bianco2019quasi}. 
Methods that apply post-capture adjustments on already rendered sRGB images~\cite{afifi2019color, afifi2020deep, afifi2022auto, serrano2025revisiting} are excluded from the comparison, as they do not operate within the camera color correction pipeline.
To enable a fair comparison in multispectral settings, we adapt the FC$^4$ and ConvMean architectures to process MS data instead of RGB inputs. 
Since all these baselines address only the first stage of the color correction pipeline (i.e., illuminant estimation), we complete the remaining steps following the traditional correction procedure described in \Cref{sec:traditional_pipeline}.
Quantitative results are summarized in \Cref{tab:main_results}, where the $\Delta$E$_{00}$ color distance and the Reproduction Error are presented with their statistical distribution over the dataset, including the mean, median, trimean, best-quartile (B-25), worst-quartile (W-25), 95th and 99th percentiles, and maximum values. Results are aggregated by camera type (mirrorless and mobile) by averaging the statistics of individual cameras, while the full per-camera results are reported in the Supplementary Material. Our framework consistently outperforms both RGB-only and MS-driven baselines across camera categories.
It is noteworthy that the proposed models achieve an average $\Delta$E$_{00}$ that is {\raisebox{1pt}{\tiny$\sim$}}50\% lower than that of competing methods, whose average performance is comparable to the worst-quartile (W-25) results of our approaches. 
Qualitative comparisons, reported in \Cref{fig:main_qualitative}, further corroborate these findings, demonstrating that our framework produces visually more accurate and perceptually consistent results across diverse scenes, illumination conditions, and cameras.

\paragraph{Experiments on Misaligned Data}
We further assess the adaptability of our proposed framework under realistic acquisition conditions. To this end, we also conduct an experiment using the misaligned dataset described in \Cref{sec:dataset}. This experiment simulates the spatial inconsistencies that typically arise in RGB–MS imaging systems due to imperfect optical alignment.
For our proposed architectures, the models are fine-tuned on the misaligned data by keeping all weights frozen except for the spectral encoder branch, which is responsible for extracting features from the MS modality. This strategy allows the network to adjust its spectral representation to the new, geometrically perturbed inputs without altering the overall color correction mapping. 
For the MS variants of ConvMean and FC$^4$, which do not include an explicit spectral encoder, we retrain the models from scratch following the same procedure described for the previous experiment. Methods relying solely on RGB inputs are unaffected and thus excluded from this experiment.
\Cref{tab:misaligned_results} summarizes the results obtained on the misaligned dataset, aggregated by camera type (mirrorless and mobile) following the same procedure described for \Cref{tab:main_results}. Full per-camera results are reported in the Supplementary Material. As shown, our proposed models exhibit only a minor decrease in accuracy relative to the aligned case, remaining the top-performing methods by a significant margin and confirming their adaptability to realistic cross-sensor misalignments. As for the multispectral illuminant estimation baselines, their performance remains comparable to the results reported in \Cref{tab:main_results}, since their architectures strongly downsample the input and thus mitigate the effect of spatial misregistration.

\paragraph{Ablation 1: Impact of Multispectral Information}
To assess the contribution of the multispectral information in our framework, we conduct an ablation study by training all models without their spectral encoder branch. This configuration effectively reduces the input to RGB-only data, allowing us to isolate the impact of the multispectral modality on overall performance.
The results, summarized in Table~\ref{tab:ablation}, highlight that our proposed modifications for integrating multispectral cues are the key factor driving the superior performance of these architectures in the color correction task, yielding improvements of up to 50\% over the RGB-only setup. This confirms that leveraging complementary spectral information substantially enhances color correction accuracy.

\paragraph{Ablation 2: Robustness to Exposure Changes}
One of the key advantages of traditional color correction pipelines lies in their inherent exposure invariance, a consequence of their linear formulation. Our approach, in contrast, relies on non-linear transformations and therefore does not possess this property by design. To evaluate the extent to which exposure variations affect performance, we simulate different exposure levels by multiplying both the input and ground-truth images by a factor $\alpha$.
We consider two reduced exposure levels, $\alpha=0.75$ and $\alpha=0.5$, and report the corresponding results in Table~\ref{tab:exposure_invariance}. For this experiment, we focus on Reproduction Error, as $\Delta$E$_{00}$ decreases at lower brightness levels due to the way it is mathematically formulated.
As expected, the Reproduction Error slightly increases at lower exposure levels, indicating a modest degradation in performance. Nonetheless, the models maintain higher accuracy than all baselines built on the traditional color correction pipeline, including the spectral variant of ConvMean~\cite{Gong2019ConvolutionalMA}, which represents the strongest baseline.

\paragraph{Limitations}
Our method aims at directly providing the best final color-corrected estimate given an input non-corrected image. However, the end-to-end nature of our approach does not rely on explicit intermediate estimates such as illuminant or reflectance. Consequently, it offers less interpretability than modular pipelines, which can provide diagnostic or reusable intermediate outputs. Nonetheless, in applications where the primary goal is achieving the highest color accuracy, our end-to-end approach remains advantageous.

\begin{table}
\caption{Impact of adding multispectral information to the proposed framework. We report mean $\Delta$E$_{00}$ and Reproduction Error obtained with (w/) and without (w/o) MS data, showing the increase of performance when spectral data is used.}
\label{tab:ablation}
\resizebox{\columnwidth}{!}{%
\begin{tabular}{llcccc}
    \toprule
                              & \multicolumn{1}{c}{}       & \multicolumn{2}{c}{Mean $\Delta$E$_{00}$ $\downarrow$} & \multicolumn{2}{c}{Mean Repr. Error $\downarrow$} \\ \cmidrule(lr){3-4} \cmidrule(lr){5-6} 
    Camera                    & \multicolumn{1}{l}{Method} & \shortstack{w/o\\MS data}     & \shortstack{w/\\MS data}   &  \shortstack{w/o\\MS data}     & \shortstack{w/\\MS data}     \\ \midrule
    \multirow{3}{*}{\vtop{\hbox{\strut Mirrorless}\hbox{\strut Sensors}}} & LPIENet            & 3.23                     & 1.74    & 4.71                     & 3.23                        \\
                              & LPIENet-small      & 3.33                     & 2.09    & 4.85                     & 3.56                        \\
                              & cmKAN-light        & 3.06                     & 1.60    & 4.40                     & 2.92                        \\ \midrule

    \multirow{3}{*}{\vtop{\hbox{\strut Mobile}\hbox{\strut Sensors}}} & LPIENet         & 3.29                     & 1.66    & 4.71                     & 3.11                        \\
                              & LPIENet-small      & 3.48                     & 1.80    & 4.84                     & 3.27                        \\
                              & cmKAN-light        & 3.03                     & 1.49    & 4.38                     & 2.84                        \\ \bottomrule
\end{tabular}%
}

\end{table}

\begin{table}
\caption{Mean Reproduction Error for different exposure levels, simulated by scaling both input and ground-truth images by a factor $\alpha$. Although Reproduction Error slightly increases for lower exposures, all models remain robust and outperform the other baselines. We report SpectralConvMean~\cite{Gong2019ConvolutionalMA} as the representative baseline, since it is the best-performing one.}
\label{tab:exposure_invariance}
\resizebox{\columnwidth}{!}{%
\begin{tabular}{llccc}
        \toprule
                                  & \multicolumn{1}{c}{}       & \multicolumn{3}{c}{Mean Reproduction Error $\downarrow$} \\ \cmidrule(lr){3-5} 
        Camera                    & \multicolumn{1}{l}{Method} & $\alpha = 1$     & $\alpha = 0.75$              & $\alpha = 0.5$           \\ \midrule
        \multirow{4}{*}{\vtop{\hbox{\strut Mirrorless}\hbox{\strut Sensors}}} 
                                  & SpectralConvMean~\cite{Gong2019ConvolutionalMA}            & 4.24                     & 4.25    & 4.30        \\ \\\noalign{\vskip -1em} \cdashline{2-5} \\\noalign{\vskip -1em}
                                  & LPIENet~(Ours)            & 3.23  & 3.41                     & 3.84                \\
                                  & LPIENet-small~(Ours)      & 3.57  & 3.70                     & 4.04                \\
                                  & cmKAN-light~(Ours)        & 2.91  & 2.98                     & 3.18                \\             \midrule

        \multirow{4}{*}{\vtop{\hbox{\strut Mobile}\hbox{\strut Sensors}}} 
                                  & SpectralConvMean~\cite{Gong2019ConvolutionalMA}            & 4.35                     & 4.35    & 4.36        \\ \\\noalign{\vskip -1em} \cdashline{2-5} \\\noalign{\vskip -1em}
                                  & LPIENet~(Ours)            & 3.11  & 3.28                     & 3.73                \\
                                  & LPIENet-small~(Ours)      & 3.27  & 3.38                     & 3.67                \\
                                  & cmKAN-light~(Ours)        & 2.84  & 2.90                     & 3.08                \\             \bottomrule
        \end{tabular}%
}

\end{table}

\section{Conclusion}
\label{sec:conclusion}
We presented a unified framework for end-to-end color correction that jointly leverages high-resolution RGB data and low-resolution multispectral information. Unlike prior works that treat the stages of the color correction pipeline independently, our approach integrates them within a single learning-based model. 
To enable training and evaluation under physically consistent conditions, we generated a new dataset by rendering hyperspectral reflectance data into realistic RGB and MS images, including a second version with spatial misalignments that replicate real-world acquisition imperfections. We demonstrated the flexibility of our framework by refactoring two state-of-the-art image-to-image architectures to operate within our dual-input setting.
Extensive experiments demonstrate that the proposed framework consistently outperforms existing RGB-only and MS-driven approaches across multiple camera models. The results confirm that jointly learning illuminant estimation, discounting, and color transformation, while exploiting multispectral cues, leads to substantial gains in both color accuracy and stability. By considering two distinct architectures and performing comprehensive ablations, we show that these improvements stem primarily from the proposed framework rather than the specific network design. This suggests that the framework is not tied to a specific architecture: any future backbone, if more powerful, can be seamlessly integrated within the same formulation to further enhance performance.
Furthermore, the proposed architectures maintain strong performance under challenging conditions such as exposure variations and spatial misalignments, validating their robustness for practical deployment. We anticipate that our framework and dataset will support further advancement in color correction research.

\section*{Acknowledgments} 
JVC acknowledges Grants PID2021-128178OB-I00 and PID2024-162555OB-I00 funded by MCIN/AEI/10.13039/ 501100011033 and by ERDF ``A way of making Europe", the Generalitat de Catalunya CERCA Program, and the 2025 Leonardo Grant for Scientific Research and Cultural Creation from the BBVA Foundation. The BBVA Foundation accepts no responsibility for the opinions, statements and contents included in the project and/or the results thereof, which are entirely the responsibility of the authors. 
This work was partially supported by the MUR under the grant ``Dipartimenti di Eccellenza 2023-2027" of the Department of Informatics, Systems and Communication of the University of Milano-Bicocca, Italy.

{
\small
\bibliographystyle{ieeenat_fullname}
\bibliography{main}}

@String(IJCV = {Int. J. Comput. Vis.})

@String(CVPR= {IEEE Conf. Comput. Vis. Pattern Recog.})

@String(ICCV= {Int. Conf. Comput. Vis.})

@String(ECCV= {Eur. Conf. Comput. Vis.})

@String(BMVC= {Brit. Mach. Vis. Conf.})

@String(TIP  = {IEEE Trans. Image Process.})

@String(ICIP = {IEEE Int. Conf. Image Process.})

@String(ICLR = {Int. Conf. Learn. Represent.})

@String(IJCV  = {IJCV})

@String(CVPR  = {CVPR})

@String(ICCV  = {ICCV})

@String(ECCV  = {ECCV})

@String(BMVC  =	{BMVC})

@String(ICIP  = {ICIP})

@String(ICLR  = {ICLR})

@inproceedings{zhou2024joint,
  title={Joint RGB-Spectral Decomposition Model Guided Image Enhancement in Mobile Photography},
  author={Zhou, Kailai and Cai, Lijing and Wang, Yibo and Zhang, Mengya and Wen, Bihan and Shen, Qiu and Cao, Xun},
  booktitle={ECCV},
  OPTpages={19--36},
  year={2024},
  OPTorganization={Springer}
}

@inproceedings{li2025multi,
  title={Multi-Spectral Image Color Reproduction},
  author={Li, Jiacheng and Chen, Chang and Hu, Xue and Song, Fenglong and Yan, Youliang and Xiong, Zhiwei},
  booktitle={WACV},
  OPTpages={8400--8409},
  year={2025},
  OPTorganization={IEEE}
}

@article{barnard2002data,
  title={A data set for color research},
  author={Kobus Barnard and Lindsay Martin and Brian V. Funt and Adam Coath},
  journal={Color Research \& Application},
  year={2002},
  volume={27},
  pages={147-151},
}

@inproceedings{van2005color,
  title={Color constancy based on the grey-edge hypothesis},
  author={Van De Weijer, Joost and Gevers, Theo},
  booktitle={ICIP},
  OPTvolume={2},
  OPTpages={II--722},
  year={2005},
  OPTorganization={IEEE}
}

@inproceedings{barron2015convolutional,
  title={Convolutional color constancy},
  author={Barron, Jonathan T},
  booktitle={ICCV},
  OPTpages={379--387},
  year={2015}
}

@inproceedings{hu2017fc4,
  title={Fc4: Fully convolutional color constancy with confidence-weighted pooling},
  author={Hu, Yuanming and Wang, Baoyuan and Lin, Stephen},
  booktitle={CVPR},
  OPTpages={4085--4094},
  year={2017}
}

@inproceedings{barron2017fast,
  title={Fast fourier color constancy},
  author={Barron, Jonathan T and Tsai, Yun-Ta},
  booktitle={CVPR},
  OPTpages={886--894},
  year={2017}
}

@inproceedings{bianco2019quasi,
  title={Quasi-unsupervised color constancy},
  author={Bianco, Simone and Cusano, Claudio},
  booktitle={CVPR},
  OPTpages={12212--12221},
  year={2019}
}

@inproceedings{Gong2019ConvolutionalMA,
  title={Convolutional Mean: A Simple Convolutional Neural Network for Illuminant Estimation},
  author={Han Gong},
  booktitle={BMVC},
  year={2019},
}

@inproceedings{bianco2015color,
  title={Color constancy using CNNs},
  author={Bianco, Simone and Cusano, Claudio and Schettini, Raimondo},
  booktitle={CVPR},
  OPTpages={81--89},
  year={2015}
}

@article{khan2017illuminant,
  title={Illuminant estimation in multispectral imaging},
  author={Khan, Haris Ahmad and Thomas, Jean-Baptiste and Hardeberg, Jon Yngve and Laligant, Olivier},
  journal={JOSA A},
  volume={34},
  number={7},
  pages={1085--1098},
  year={2017},
  publisher={Optical Society of America}
}

@article{erba2024rgb,
  title={RGB color constancy using multispectral pixel information},
  author={Erba, Ilaria and Buzzelli, Marco and Schettini, Raimondo},
  journal={JOSA A},
  volume={41},
  number={2},
  pages={185--194},
  year={2024},
  publisher={Optica Publishing Group}
}

@article{koskinen2024single,
  title={Single pixel spectral color constancy},
  author={Koskinen, Samu and Acar, Erman and K{\"a}m{\"a}r{\"a}inen, Joni-Kristian},
  journal={IJCV},
  volume={132},
  number={2},
  pages={287--299},
  year={2024},
  publisher={Springer}
}

@article{Erba2024ImprovingRI,
  title={Improving RGB illuminant estimation exploiting spectral average radiance.},
  author={Ilaria Erba and Marco Buzzelli and Jean-Baptiste Thomas and Jon Y. Hardeberg and Raimondo Schettini},
  journal={JOSA A},
  year={2024},
  volume={41 3},
  pages={516-526},
}

@inproceedings{li2021multispectral,
  title={Multispectral illumination estimation using deep unrolling network},
  author={Li, Yuqi and Fu, Qiang and Heidrich, Wolfgang},
  booktitle={ICCV},
  OPTpages={2672--2681},
  year={2021}
}

@inproceedings{glatt2024beyond,
  title={Beyond RGB: a real world dataset for multispectral imaging in mobile devices},
  author={Glatt, Ortal and Ater, Yotam and Kim, Woo-Shik and Werman, Shira and Berby, Oded and Zini, Yael and Zelinger, Shay and Lee, Sangyoon and Choi, Heejin and Soloveichik, Evgeny},
  booktitle={WACV},
  OPTpages={4344--4354},
  year={2024}
}

@book{fairchild2013color,
  title={Color appearance models},
  author={Fairchild, Mark D},
  year={2013},
  publisher={John Wiley \& Sons}
}

@inproceedings{cheng2015beyond,
  title={Beyond white: Ground truth colors for color constancy correction},
  author={Cheng, Dongliang and Price, Brian and Cohen, Scott and Brown, Michael S},
  booktitle={ICCV},
  OPTpages={298--306},
  year={2015}
}

@inproceedings{finlayson1993diagonal,
  title={Diagonal transforms suffice for color constancy},
  author={Finlayson, Graham D and Drew, Mark S and Funt, Brian V},
  booktitle={ICCV},
  OPTpages={164--171},
  year={1993},
  OPTorganization={IEEE}
}

@article{mccamy1976color,
  title={A color-rendition chart},
  author={McCamy, Calvin S and Marcus, Harold and Davidson, James G and others},
  journal={Journal of Applied Photographic Engineering},
  volume={2},
  number={3},
  pages={95--99},
  year={1976}
}

@misc{CIE19,
  title = {Colour-matching functions of CIE 1931 standard colorimetric observer},
  howpublished = {\url{https://cie.co.at/datatable/cie-1931-colour-matching-functions-2-degree-observer}},
  DOI = {10.25039/cie.ds.xvudnb9b},
  author = {International Commission on Illumination (CIE)},
  year = {2019}
}

@inproceedings{conde2023perceptual,
  title={Perceptual image enhancement for smartphone real-time applications},
  author={Conde, Marcos V and Vasluianu, Florin and Vazquez-Corral, Javier and Timofte, Radu},
  booktitle={WACV},
  OPTpages={1848--1858},
  year={2023}
}

@inproceedings{nikonorov2025color,
  title={Color Matching Using Hypernetwork-Based Kolmogorov-Arnold Networks},
  author={Nikonorov, Artem and Perevozchikov, Georgy and Korepanov, Andrei and Mehta, Nancy and Afifi, Mahmoud and Ershov, Egor and Timofte, Radu},
  booktitle={ICCV},
  OPTpages={...},
  year={2025}
}

@article{howard2017mobilenets,
  title={Mobilenets: Efficient convolutional neural networks for mobile vision applications},
  author={Howard, Andrew G and Zhu, Menglong and Chen, Bo and Kalenichenko, Dmitry and Wang, Weijun and Weyand, Tobias and Andreetto, Marco and Adam, Hartwig},
  journal={arXiv preprint arXiv:1704.04861},
  year={2017}
}

@inproceedings{woo2018cbam,
  title={Cbam: Convolutional block attention module},
  author={Woo, Sanghyun and Park, Jongchan and Lee, Joon-Young and Kweon, In So},
  booktitle={ECCV},
  OPTpages={3--19},
  year={2018}
}

@inproceedings{ronneberger2015u,
  title={U-net: Convolutional networks for biomedical image segmentation},
  author={Ronneberger, Olaf and Fischer, Philipp and Brox, Thomas},
  booktitle={MICCAI},
  OPTpages={234--241},
  year={2015},
  OPTorganization={Springer}
}

@inproceedings{liu2024kan,
  title={Kan: Kolmogorov-arnold networks},
  author={Liu, Ziming and Wang, Yixuan and Vaidya, Sachin and Ruehle, Fabian and Halverson, James and Solja{\v{c}}i{\'c}, Marin and Hou, Thomas Y and Tegmark, Max},
  booktitle={ICLR},
  year={2025}
}

@inproceedings{gehler2008bayesian,
  title={Bayesian color constancy revisited},
  author={Gehler, Peter Vincent and Rother, Carsten and Blake, Andrew and Minka, Tom and Sharp, Toby},
  booktitle={CVPR},
  OPTpages={1--8},
  year={2008},
  OPTorganization={IEEE}
}

@article{cheng2014illuminant,
  title={Illuminant estimation for color constancy: why spatial-domain methods work and the role of the color distribution},
  author={Cheng, Dongliang and Prasad, Dilip K and Brown, Michael S},
  journal={JOSA A},
  volume={31},
  number={5},
  pages={1049--1058},
  year={2014},
  publisher={Optical Society of America}
}

@article{ershov2020cube++,
  title={The cube++ illumination estimation dataset},
  author={Ershov, Egor and Savchik, Alexey and Semenkov, Illya and Bani{\'c}, Nikola and Belokopytov, Alexander and Senshina, Daria and Ko{\v{s}}{\v{c}}evi{\'c}, Karlo and Suba{\v{s}}i{\'c}, Marko and Lon{\v{c}}ari{\'c}, Sven},
  journal={IEEE Access},
  volume={8},
  pages={227511--227527},
  year={2020},
  publisher={IEEE}
}

@inproceedings{du2025automatic,
  title={Automatic Spectral Calibration of Hyperspectral Images: Method, Dataset and Benchmark},
  author={Du, Zhuoran and You, Shaodi and Cheng, Cheng and Wei, Shikui},
  booktitle={CVPR},
  OPTpages={28081--28090},
  year={2025}
}

@article{thomas2025trends,
  title={Trends in snapshot spectral imaging: systems, processing, and quality},
  author={Thomas, Jean-Baptiste and Lapray, Pierre-Jean and Le Moan, Steven},
  journal={Sensors},
  volume={25},
  number={3},
  pages={675},
  year={2025},
  publisher={MDPI}
}

@article{lapray2014multispectral,
  title={Multispectral filter arrays: Recent advances and practical implementation},
  author={Lapray, Pierre-Jean and Wang, Xingbo and Thomas, Jean-Baptiste and Gouton, Pierre},
  journal={Sensors},
  volume={14},
  number={11},
  pages={21626--21659},
  year={2014},
  publisher={MDPI}
}

@article{su2021multi,
  title={Multi-spectral fusion and denoising of color and near-infrared images using multi-scale wavelet analysis},
  author={Su, Haonan and Jung, Cheolkon and Yu, Long},
  journal={Sensors},
  volume={21},
  number={11},
  pages={3610},
  year={2021},
  publisher={MDPI}
}

@inproceedings{tian2023enhancing,
  title={Enhancing Low-Light Images Using Infrared Encoded Images},
  author={Tian, Shulin and Wang, Yufei and Wan, Renjie and Yang, Wenhan and Kot, Alex C and Wen, Bihan},
  booktitle={ICIP},
  OPTpages={465--469},
  year={2023},
  OPTorganization={IEEE}
}

@article{ISO22028-2:2013,
  title={22028-2:2013 Photography and graphic technology — Extended colour encodings for digital image storage, manipulation and interchange-Part 2: Reference output medium metric RGB colour image encoding (ROMM RGB)},
  author={ISO},
  journal={International Organization for Standardization},
  year={2013}
}

@misc{spectricity,
    author = {{S}pectricity | {M}ultispectral {I}maging {S}olutions},
    howpublished = {\url{https://spectricity.com/}},
    note = {[Accessed 17-11-2025]}, }

@inproceedings{kingma2014adam,
  title={Adam: A method for stochastic optimization},
  author={Kingma, Diederik P},
  booktitle={ICLR},
  year={2015}
}

@article{finlayson2016reproduction,
  title={The reproduction angular error for evaluating the performance of illuminant estimation algorithms},
  author={Finlayson, Graham D and Zakizadeh, Roshanak and Gijsenij, Arjan},
  journal={IEEE TPAMI},
  volume={39},
  number={7},
  pages={1482--1488},
  year={2016},
  publisher={IEEE}
}

@article{robertson1977cie,
  title={The CIE 1976 color-difference formulae},
  author={Robertson, Alan R},
  journal={Color Research \& Application},
  volume={2},
  number={1},
  pages={7--11},
  year={1977},
  publisher={Wiley Online Library}
}

@article{luo2001development,
  title={The development of the CIE 2000 colour-difference formula: CIEDE2000},
  author={Luo, M Ronnier and Cui, Guihua and Rigg, Bryan},
  journal={Color Research \& Application},
  volume={26},
  number={5},
  pages={340--350},
  year={2001},
  publisher={Wiley Online Library}
}

@inproceedings{loshchilov2016sgdr,
  title={Sgdr: Stochastic gradient descent with warm restarts},
  author={Loshchilov, Ilya and Hutter, Frank},
  booktitle={ICLR},
  year={2017}
}

@techreport{vagni2007survey,
  title={Survey of Hyperspectral and Multispectral Imaging Technologies (Etude sur les technologies d'imagerie hyperspectrale et multispectrale)},
  author={Vagni, Fabrizio},
  year={2007}
}

@inproceedings{ignatov2020replacing,
  title={Replacing mobile camera isp with a single deep learning model},
  author={Ignatov, Andrey and Van Gool, Luc and Timofte, Radu},
  booktitle={CVPR Workshops},
  OPTpages={536--537},
  year={2020}
}

@inproceedings{afifi2019color,
  title={When color constancy goes wrong: Correcting improperly white-balanced images},
  author={Afifi, Mahmoud and Price, Brian and Cohen, Scott and Brown, Michael S},
  booktitle={CVPR},
  OPTpages={1535--1544},
  year={2019}
}

@inproceedings{afifi2020deep,
  title={Deep white-balance editing},
  author={Afifi, Mahmoud and Brown, Michael S},
  booktitle={CVPR},
  OPTpages={1397--1406},
  year={2020}
}

@inproceedings{serrano2025revisiting,
  title={Revisiting Image Fusion for Multi-Illuminant White-Balance Correction},
  author={Serrano-Lozano, David and Arora, Aditya and Herranz, Luis and Derpanis, Konstantinos G and Brown, Michael S and Vazquez-Corral, Javier},
  booktitle={ICCV},
  year={2025}
}

@inproceedings{afifi2022auto,
  title={Auto white-balance correction for mixed-illuminant scenes},
  author={Afifi, Mahmoud and Brubaker, Marcus A and Brown, Michael S},
  booktitle={WACV},
  OPTpages={1210--1219},
  year={2022}
}

@inproceedings{afifi2025time,
  title={Time-aware auto white balance in mobile photography},
  author={Afifi, Mahmoud and Zhao, Luxi and Punnappurath, Abhijith and Abdelsalam, Mohamed A and Zhang, Ran and Brown, Michael S},
  booktitle={ICCV},
  OPTpages={5038--5047},
  year={2025}
}

@article{finlayson2015color,
  title={Color correction using root-polynomial regression},
  author={Finlayson, Graham D and Mackiewicz, Michal and Hurlbert, Anya},
  journal={IEEE TIP},
  volume={24},
  number={5},
  pages={1460--1470},
  year={2015},
  publisher={IEEE}
}

@article{hong2001study,
  title={A study of digital camera colorimetric characterization based on polynomial modeling},
  author={Hong, Guowei and Luo, M Ronnier and Rhodes, Peter A},
  journal={Color Research \& Application},
  volume={26},
  number={1},
  pages={76--84},
  year={2001},
  publisher={Wiley Online Library}
}

\end{document}